\documentclass[11pt]{article}

\usepackage{acl}

\usepackage{fontspec}
\usepackage{polyglossia}
\setmainlanguage{english}
\setotherlanguage{arabic}
\newfontfamily\arabicfont[   Script=Arabic,   Scale=1.2,   BoldFont=Amiri-Bold.ttf,   ItalicFont=Amiri-Italic.ttf,   BoldItalicFont=Amiri-BoldItalic.ttf ]{Amiri-Regular.ttf}

\usepackage{caption}
\usepackage{latexsym}
\usepackage{stfloats}
\usepackage{multirow}
\usepackage{cuted}
\usepackage{array}
\usepackage{microtype}
\usepackage{inconsolata}
\usepackage{booktabs}
\usepackage{amsmath}
\usepackage{makecell}
\usepackage{tabularx}
\usepackage{amssymb}
\usepackage{lineno}
\usepackage{enumitem}
\usepackage{hyperref}
\usepackage{float}
\usepackage{graphicx}

\newcommand{\cmark}{\ensuremath{\checkmark}}
\newcommand{\xmark}{\ensuremath{\times}}

\title{Shaer: Controlled Arabic Poetry Generation with Meter Subform and Semantic Conditioning}

\usepackage{tikz}
\usetikzlibrary{positioning,arrows.meta,shapes.geometric}
\author{
\mbox{Ahmad Abbas \qquad
Tamara Fakih \qquad
Nour Fakih \qquad
Ammar Mohanna} \\[4pt]
Department of Electrical and Computer Engineering \\
American University of Beirut, Beirut, Lebanon \\
\texttt{\{asa108,tmf14,nhf16\}@mail.aub.edu}
\qquad
\texttt{am288@aub.edu.lb} \\[4pt]
}

\newfontfamily\arabicfonttt[   Script=Arabic,   Scale=1.1 ]{Amiri-Regular.ttf}

\newcommand{\ArCell}[1]{%
\begin{minipage}[t]{\linewidth}
\begin{Arabic}
#1
\end{Arabic}
\end{minipage}
}

\begin{document}

\maketitle

\begin{figure*}[t]
    \centering
    \includegraphics[width=\textwidth]{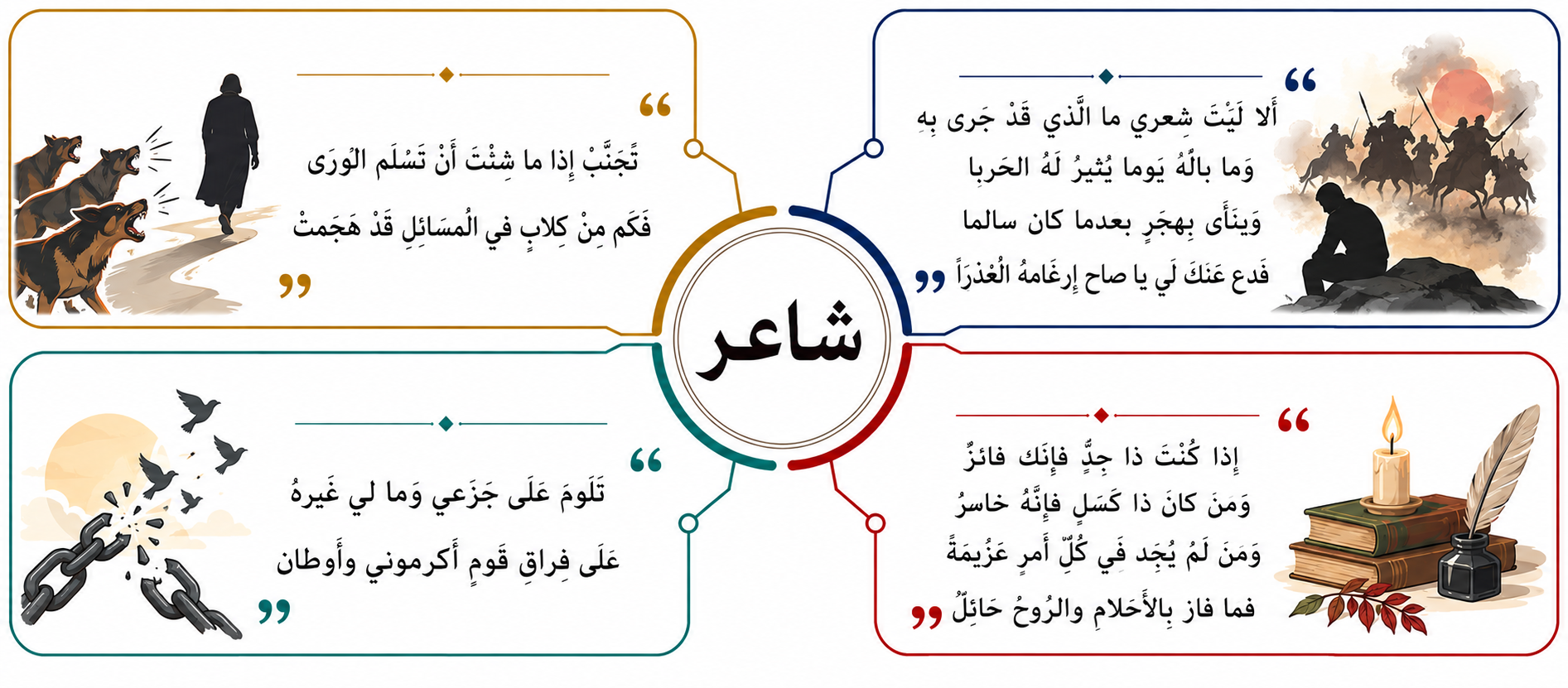}
    \caption{Example poems generated by Shaer. Each poem is generated from a semantic description, meter specification, and length constraint derived from a held-out test example.  }\label{fig:intro_poetry_examples}
\end{figure*}

\begin{abstract}
Classical Arabic poetry generation requires simultaneously satisfying semantic, linguistic, and fine-grained prosodic constraints. Existing systems typically control broad poetic attributes but do not jointly model semantic intent, meter subform, and poem length. We present \textsc{Shaer}, a controllable Classical Arabic poetry generation framework jointly conditioned on natural-language descriptions, meter subforms, and target hemistich counts. To support this task, we construct an enriched corpus of 116,032 classical Arabic poems derived from Ashaar, containing normalized meter-subform labels and automatically generated, validated semantic descriptions. We then adapt Yehia-7B using QLoRA-based supervised fine-tuning with a completion-only objective. Our evaluation combines automatic assessment of base-meter conformity, requested-subform adherence, and length control with three LLM judges, blinded human evaluation, and memorization analysis. \textsc{Shaer} achieves 95.17\% base-meter accuracy, 91.75\% poem-level meter-subform accuracy, and 83.40\% exact count accuracy. Relative to its untuned foundation model, these results represent gains of 68.68, 57.77, and 38.93 percentage points, respectively; \textsc{Shaer} also attains the highest base-meter accuracy among all evaluated systems. Multi-LLM evaluation and a blinded human assessment of top-ranked outputs further indicate competitive semantic and literary quality. Finally, analysis of all 3,481 test generations finds no exact copies from the training corpus or paired source poems. Code, models, and datasets are publicly available.\footnote{Code: \url{https://github.com/AhmaddAbbass/Shaer}. Models and datasets: \url{https://huggingface.co/Shaer-AI}.}
\end{abstract}

\section{Introduction}

Classical Arabic poetry is a central component of Arabic literary heritage, making controlled generation both a technically challenging problem and a culturally grounded Arabic NLP task. Unlike conventional language generation tasks, successful poetry generation must simultaneously satisfy semantic, stylistic, and prosodic constraints. Beyond producing fluent and coherent text, a generated poem must adhere to a prescribed meter while faithfully expressing the intended meaning. These requirements make Arabic poetry generation a particularly demanding controllable generation problem. A brief overview of Classical Arabic poetic structure, meter, and meter subforms is provided in Appendix~\ref{app:poetry-background}.

Despite recent progress in Arabic language generation and computational poetry, fine-grained control over generated Arabic poetry remains challenging. Existing generation settings largely operate over broad structural or categorical attributes, leaving important dimensions of poetic control insufficiently addressed. In particular, jointly controlling natural-language semantic intent, fine-grained meter subforms, and explicit poem length remains underexplored. This limits the ability to generate poems that simultaneously realize a desired meaning and satisfy precise prosodic and structural requirements.

To address these gaps, we introduce \textsc{Shaer}, a controllable Arabic poetry generation framework that jointly models semantic intent and poetic structure. By conditioning generation on natural-language descriptions, meter subforms, and target hemistich counts, \textsc{Shaer} enables fine-grained control over meaning, meter, and poem structure. Figure~\ref{fig:intro_poetry_examples} presents representative poems generated by \textsc{Shaer} from prompts drawn from the test split of the released dataset.

Our contributions are as follows: (1) We introduce a controllable Arabic poetry generation task conditioned jointly on semantic descriptions, meter subforms, and requested poem length, enabling fine-grained control over semantic and structural properties. (2) We construct an enriched Arabic poetry corpus derived from Ashaar, comprising 116,032 classical poems annotated with normalized meter subforms and semantic descriptions, together with the complete data processing pipeline. (3) We develop \textsc{Shaer}, a controllable Arabic poetry generation model built by adapting Yehia-7B, a strong Arabic-focused language model, through parameter-efficient fine-tuning and explicit conditioning on semantic and prosodic constraints. (4) We introduce a multi-perspective evaluation framework combining automatic assessment of base-meter conformity, meter-subform adherence, and length control with three LLM judges, blinded human evaluation, and memorization and copying analysis.

Experimental results show that \textsc{Shaer} attains the highest base-meter accuracy among all evaluated systems and achieves high requested-subform adherence. Relative to its untuned foundation model, Yehia-7B, \textsc{Shaer} provides markedly stronger base-meter, meter-subform, and length control. Evaluation with three LLM judges and a blinded human assessment of top-ranked outputs further indicates that these gains in controllability are achieved while retaining competitive semantic and literary quality.

\section{Related Work}  
\label{sec:related-work}

\subsection{Arabic Poetry Generation}

Early neural approaches to Arabic poetry generation relied on task-specific architectures. \citet{talafha2019hierarchical} combined keyword-based semantic guidance with Bi-GRU generation and hierarchical attention over previously generated verses, while their later work incorporated phonetic CNN subword embeddings and keyword expansion to improve metrical conformity and semantic consistency \cite{talafha2021phonetic}.

Pretrained language models subsequently enabled transfer-based approaches. \citet{beheitt2022automatic} pretrained GPT-2 on Arabic text and fine-tuned it for poetry generation, while \citet{abboushi2023fluent} fine-tuned AraGPT2 on a large poetry corpus with explicit meter and rhyme conditioning.

Recent work has expanded both resources and controllability. Ashaar \cite{alyafeai-etal-2023-ashaar} introduced datasets and pretrained models for Arabic poetry analysis and conditional generation. Fanar-Diwan \cite{fanar2026fanar20} supports controllable classical Arabic poetry generation conditioned on meter, topic, era, poet, and rhyme. From an instruction-following perspective, \citet{sadallah-etal-2026-instruction} introduced a large-scale resource spanning Modern Standard Arabic and multiple dialects and supporting poetry generation, continuation, revision, and analysis under user-specified constraints.

Building on this progression, \textsc{Shaer} offers a complementary approach to controllable classical Arabic poetry generation through joint conditioning on natural-language semantic descriptions, meter subforms, and explicit poem length.

\subsection{Arabic Poetry Analysis and Evaluation}

Arabic poetry analysis has extensively addressed meter identification. \citet{alshaibani2020meter} proposed a character-based bidirectional recurrent model for classifying poetry into 14 meters without traditional Arudi representation, while \citet{ABANDAH2022377} developed recurrent models covering the 16 classical meters and prose, alongside automatic diacritization. We reproduce the latter meter-classification architecture to assess base-meter conformity.

For finer-grained prosody, \citet{qarah-2024-arapoembert} introduced AraPoemBERT, a poetry-specific language model evaluated on several analysis tasks, including meter-subform classification. We reproduce this setting to assess subform adherence in controllable generation.

At the semantic level, \citet{al-ghallabi-etal-2025-fann} introduced Fann or Flop, a benchmark for LLM understanding of Arabic poetry across semantic, figurative, prosodic, and cultural dimensions. Our evaluation similarly adopts a multidimensional perspective, combining automatic assessment of meter, subform, and length control with LLM-based and human judgments of semantic and literary quality.

\section{Dataset}
\label{sec:data}

\subsection{Source Corpus}

Our work builds on Ashaar \cite{alyafeai-etal-2023-ashaar}, a large-scale Arabic poetry resource containing 254,630 poems from multiple online repositories. Ashaar supports several poetry-related tasks and provides poem-level metadata including meter, theme, poet identity, era, and descriptions. However, its metadata organization and description coverage are not directly suited to controllable generation based on fine-grained poetic structure and semantic intent. Appendix~\ref{app:ashaar-analysis} provides a detailed analysis of the original corpus.

\subsection{Corpus Construction}

We construct an enriched corpus for controllable Arabic poetry generation through multi-stage filtering and normalization. We retain only classical Arabic poetry in the canonical Khalilian meters, removing colloquial or non-classical forms, malformed poems, and entries with invalid verse structure. To align with the 1,024-token maximum sequence length used for supervised fine-tuning, we exclude poems exceeding twenty verses. A post-tokenization audit using the final instruction format found that only 4 of 109,070 training sequences (0.0037\%) exceeded this limit, by 5--25 tokens, with none in the validation or test sets. Finally, we remove insufficiently represented meter categories. Appendix~\ref{app:data-cleaning} details the cleaning, normalization, and filtering procedures.



\subsection{Meter Subform Annotation and Validation}

All 116,032 retained poems inherit their base-meter labels from Ashaar. Explicit meter-subform metadata was available for 5,106 poems (4.4\%); we normalized and retained these labels without independent re-annotation. The remaining 110,926 poems (95.6\%) were assigned the default \textarabic{تام} (Complete) label, corresponding to the canonical realization of their base meter in classical Arabic prosody.

We assessed this default assignment in two ways. First, an author formally trained in Arabic prosody manually inspected 1,000 default-labeled poems, confirming the Complete form in 97.2\% of cases. Second, we reproduced the AraPoemBERT meter-subform classification setting \cite{qarah-2024-arapoembert}, achieving 97.38\% accuracy compared with the originally reported 97.79\%. Applied to all 110,926 default-labeled poems, the classifier predicted Complete for 103,748 (93.53\%). Agreement exceeded 98.7\% for nine of thirteen base meters, with disagreements concentrated in four meters affected by classifier limitations in class coverage or reported per-class performance. We use these predictions only to assess the default-assignment convention, not to replace corpus labels; complete per-meter results and discussion are provided in Appendix~\ref{app:subform-validation}.

After normalization and default assignment, all poems have meter-subform annotations, yielding 21 distinct meter--subform combinations whose distribution is reported in Appendix~\ref{app:subform-distribution}.

\subsection{Semantic Description Generation}

Ashaar descriptions are missing or insufficiently informative for much of the corpus, limiting their use as semantic conditioning signals (Appendix~\ref{app:ashaar-analysis}). We therefore generate a semantic description for every poem.

We compare three generation configurations differing in model and prompting strategy. For each, three native Arabic-speaking annotators independently evaluate a stratified sample of 600 descriptions using a predefined Pass/Fail rubric. The best-performing configuration, based on Qwen3.5-35B-A3B \cite{qwen-2026-qwen35}, achieves a 92.8\% pass rate and is selected for corpus-wide generation. Configuration and human-evaluation details are provided in Appendix~\ref{app:description-configuration-selection}.

The selected configuration uses deterministic decoding to generate concise descriptions capturing each poem's meaning, imagery, and tone while excluding references to meter, rhyme, or length. The resulting descriptions undergo automatic structural validation and corpus-wide quality checks. Appendices~\ref{app:description-prompting} and~\ref{app:description-validation} provide full prompting, decoding, and verification details.

\subsection{Final Dataset Statistics}

The final corpus pairs each poem with its base meter, meter subform, target hemistich count, and generated semantic description, jointly supporting control over poetic form and semantic content. It comprises 116,032 poems spanning 13 classical meters and 21 meter--subform combinations, providing substantially finer-grained metrical supervision than the coarse meter categories commonly used in prior Arabic poetry generation systems. Appendix~\ref{app:final-dataset} summarizes the dataset characteristics, while Table~\ref{tab:dataset_example} illustrates the enriched metadata and controlled generation format.

\begin{table*}[t]
\centering
\small
\setlength{\tabcolsep}{4pt}
\renewcommand{\arraystretch}{1.15}

\begin{tabular}{
    >{\centering\arraybackslash}p{0.07\textwidth}
    >{\centering\arraybackslash}p{0.07\textwidth}
    >{\centering\arraybackslash}p{0.10\textwidth}
    >{\raggedleft\arraybackslash}p{0.31\textwidth}
    >{\raggedleft\arraybackslash}p{0.36\textwidth}
}
\toprule
\textbf{Meter} &
\textbf{Subform} &
\textbf{Requested Hemistichs} &
\textbf{Description} &
\textbf{Poem Hemistichs} \\
\midrule

\begin{Arabic}الوافر\end{Arabic}
&
\begin{Arabic}تام\end{Arabic}
&
4
&
\begin{Arabic}
القصيدة تتحدث عن ليلٍ طويلٍ كليل يوم القيامة، جعل السهر مستحيلاً، لدرجة أن الشاعر كاد يظن في سهادته أن الله لم يخلق نهارًا.
\end{Arabic}
&
\ArCell{%
وَلَيل مثل يَوم الحَشر طولاً\par
مَنَعَت بِهِ مِن السَهر القَرارا\par
وَكدت أَظُنُّ فيهِ مِن سُهادي\par
بِأَن اللَه لَم يَخلق نَهارا
}
\tabularnewline

\bottomrule
\end{tabular}

\caption{Example Shaer dataset instance with its conditioning signals: base meter, meter subform, target hemistich count, and semantic description.}
\label{tab:dataset_example}
\end{table*}

\subsection{Data Splits}

We partition the corpus into train (94\%), validation (3\%), and test (3\%) splits, stratified by a composite key of base meter, meter subform, and length bucket to preserve poetic-structure distributions across partitions. Length buckets are defined by poem length as 1--3, 4--6, 7--10, and 11--20 verses.

\section{Methodology}
\label{sec:method}

\subsection{Task Formulation}

We formulate Arabic poetry generation as a controllable generation task. Given a conditioning tuple

\[
x = (m, s, n, d),
\]

where $m$ is the base meter, $s$ the meter subform, $n$ the requested hemistich count, and $d$ a natural-language description of the desired content, the model generates

\[
y = (y_1, \ldots, y_T),
\]

where $y_t$ is the $t$-th generated token and $T$ the total number of generated tokens. The objective is to model $P(y \mid x)$ such that the generated poem satisfies the specified semantic, prosodic, and length constraints.

\subsection{Base Model Selection}

Model selection is informed by AraLingBench \cite{zbib2025aralingbench}, a human-annotated benchmark of Arabic linguistic competence across grammar, morphology, spelling, syntax, and reading comprehension, capabilities closely aligned with classical Arabic poetry generation.

Yehia-7B \cite{yehia2025}, a 7B-parameter Arabic-focused decoder-only language model, achieved the strongest overall performance among the evaluated models. We therefore select \texttt{Navid-AI/Yehia-7B-preview}\footnote{Hugging Face model repository handle.} as the foundation model for \textsc{Shaer}. Its Apache-2.0 license also facilitates public release of adapters and derived resources.

\subsection{Supervised Fine-Tuning}

We adapt Yehia-7B through supervised fine-tuning on the enriched corpus training split using a completion-only causal language modeling objective. Given prompt $x$ and target completion $y$, the loss is computed only over poem tokens:

\begin{equation}
\mathcal{L}(\theta)
=
-\sum_{t \in \mathcal{C}}
\log p_{\theta}(y_t \mid x, y_{<t}),
\end{equation}

where $\mathcal{C}$ denotes completion-token positions; prompt tokens are masked from the optimization objective.

To mitigate dataset imbalance, we use weighted sampling over groups defined by (base meter, meter subform, length bucket):

\[
w(g) \propto \mathrm{count}(g)^{-0.6}.
\]

The exponent $0.6$ provides moderate rebalancing between unweighted ($0$) and full inverse-frequency ($1$) sampling, increasing representation of rare groups without excessive oversampling.

Fine-tuning uses QLoRA \cite{dettmers-etal-2023-qlora}, with the base model loaded in 4-bit NF4 precision, double quantization, and bfloat16 computation. LoRA adapters are applied to all linear layers with rank $r=64$, $\alpha=128$, dropout $0.05$, and RS-LoRA enabled.

Appendix~\ref{app:sft-details} provides complete training hyperparameters, runtime settings, and instruction templates.

\section{Experimental Setup}
\label{sec:experimental_setup}

\subsection{Baselines}

We evaluate \textsc{Shaer} against three Arabic poetry generation baselines spanning foundation and poetry-specific models:

\begin{itemize}
\item \textbf{Yehia-7B} \cite{yehia2025}, the unfine-tuned foundation model underlying \textsc{Shaer}.
\item \textbf{Ashaar} \cite{alyafeai-etal-2023-ashaar}, a poetry generation model trained on the Ashaar corpus.
\item \textbf{Fanar-Diwan} \cite{fanar2026fanar20}, a classical Arabic poetry model optimized for prosodic and stylistic constraints.
\end{itemize}

Yehia-7B provides the cleanest controlled comparison, sharing \textsc{Shaer}'s architecture, tokenizer, and instruction format but lacking task-specific fine-tuning. Ashaar and Fanar-Diwan serve as system-level comparisons under their native conditioning interfaces, which do not support \textsc{Shaer}'s full set of semantic, subform, and length controls. Earlier systems in Section~\ref{sec:related-work} are not directly evaluated due to differing generation and evaluation settings and, where applicable, unavailable public checkpoints or inference implementations.

All models are evaluated on the 3,481-instance held-out test split. Yehia-7B uses the same instruction-style prompt as \textsc{Shaer}, while Ashaar and Fanar-Diwan use their native conditioning formats. Table~\ref{tab:conditioning_capabilities} summarizes conditioning capabilities; Appendices~\ref{app:native_input_representation} and~\ref{app:generation-decoding} detail input representations and decoding settings, respectively.

\begin{table*}[t]
\centering
\small
\begin{tabular}{lccccc}
\toprule
\textbf{Model} & \textbf{Meter} & \textbf{Subform} & \textbf{Rhyme} & \textbf{Hemistich Count} & \textbf{Description} \\
\midrule
Yehia-7B & \cmark$^{\dagger}$ & \cmark$^{\dagger}$ & \xmark & \cmark$^{\dagger}$ & \cmark$^{\dagger}$ \\
Ashaar & \cmark & \xmark & \cmark & \xmark & \xmark \\
Fanar-Diwan & \cmark & \xmark & \cmark & \xmark & \xmark \\
\textsc{Shaer} & \cmark & \cmark & \xmark & \cmark & \cmark \\
\bottomrule
\end{tabular}
\caption{
Conditioning capabilities of the evaluated models. $\dagger$ denotes variables supplied through prompting rather than learned as native conditioning fields.
}
\label{tab:conditioning_capabilities}
\end{table*}

\subsection{Evaluation Dimensions}

Because Arabic poetry quality cannot be captured by a single score, we evaluate generated poems using complementary structural and semantic-literary metrics. Before structural evaluation, outputs are deterministically normalized and segmented into hemistichs and bayts by pairing consecutive hemistichs; any incomplete final hemistich is excluded from bayt-level evaluation. Appendix~\ref{app:generation-parsing} provides the full normalization and parsing rules.

\paragraph{Automatic Meter Evaluation.}

We evaluate meter using an external Arabic meter classifier based on the deep bidirectional recurrent architecture of \citet{ABANDAH2022377}. Operating at the verse level, it produces a meter prediction and probability distribution over the supported classes. Appendix~\ref{app:meter-classifier} provides complete implementation and evaluation details.

We report two complementary metrics. \textit{Meter Accuracy} measures exact adherence to the requested meter by majority-voting verse-level predictions into a poem-level prediction and computing the percentage matching the target meter.

\textit{Meter Conformity} provides a softer confidence-based measure of metrical adherence. Let $m^*$ denote the requested meter and $p_i(m^*)$ its predicted probability for verse $i$. Because meter is a poem-level property, we aggregate verse-level probabilities using the geometric mean:

\begin{equation}
C_{\mathrm{poem}}
=
\left(
\prod_{i=1}^{n}
p_i(m^*)
\right)^{1/n},
\end{equation}

where $n$ is the number of verses. The geometric mean penalizes metrically inconsistent verses, providing a more faithful estimate of poem-level conformity than an arithmetic average. Final Meter Conformity is the arithmetic mean across all generated poems.

\paragraph{Automatic Meter Subform Evaluation.}

We evaluate fine-grained prosodic adherence using our reproduced AraPoemBERT 25-class meter-subform classifier \cite{qarah-2024-arapoembert}, which operates at the bayt level. We report \textit{Bayt-Level Subform Accuracy} and \textit{Poem-Level Subform Accuracy}; the latter averages class-probability vectors across valid bayts and selects the highest-scoring class. The evaluator covers 3,379 of 3,481 test targets (97.07\%), excluding targets outside its class inventory. Appendix~\ref{app:meter-subform-classifier} details classifier reconstruction, input representation, sequence-length configuration, and unsupported meter-subform classes.

For \textsc{Shaer} and Yehia-7B, these metrics measure adherence to the requested subform. Because Ashaar and Fanar-Diwan do not support subform conditioning, their outputs are evaluated against the corresponding source poem's subform, measuring reference-subform agreement rather than controllability.

\paragraph{Hemistich Count Evaluation.}

The second structural dimension evaluates adherence to the requested hemistich count using two complementary metrics.

\textit{Count Accuracy} measures exact count matching as the percentage of generated poems whose hemistich count equals the requested count.

\textit{Count Adherence} (CA) provides a softer measure of length conformity. Let $H_g$ and $H_r > 0$ denote the generated and requested hemistich counts, respectively:

\begin{equation}
\mathrm{CA}
=
\max\left(
0,\,
1 -
\frac{|H_g - H_r|}{H_r}
\right).
\end{equation}

The score equals 1 for an exact match and decreases with relative count discrepancy, reaching 0 when the discrepancy is at least as large as the requested count. Normalization by $H_r$ makes the metric more sensitive to absolute deviations for shorter requested poems. Final Count Adherence is the arithmetic mean across all generated poems.

Both metrics are reported only for models supporting hemistich-count conditioning, namely \textsc{Shaer} and Yehia-7B.

\paragraph{LLM-Based Evaluation.}

Semantic and literary quality are evaluated using three LLM judges: Qwen3-235B-A22B \cite{qwen3technicalreport}, Qwen 3.7 Max \cite{qwen37max}, and GPT-5.6 Terra \cite{openai2026gpt56}. Qwen3-235B-A22B was applied to all 3,481 held-out outputs from each system, while Qwen 3.7 Max and GPT-5.6 Terra evaluate a shared subset of 2,500 test instances across the four systems. Following recent work on LLM-based evaluation of open-ended generation, we assess five dimensions: \textit{Description Adherence}, \textit{Meaning}, \textit{Fluency}, \textit{Coherence}, and \textit{Poeticness}. Table~\ref{tab:eval_metrics} defines the evaluation dimensions.

All three judges use identical dimension-specific rubrics. Each dimension is assessed independently on a strict 1--5 integer scale, where higher scores indicate better performance. Description Adherence is evaluated only for description-conditioned models, namely \textsc{Shaer} and Yehia-7B. Scores are first averaged over the available valid responses separately for each judge and system. The equal-weight mean of the three judge-level means is reported in the main results. To reduce evaluation variance, all judgments use deterministic decoding (\texttt{temperature}=0). Full evaluation rubrics, decoding settings, coverage details, and judge-specific results are provided in Appendix~\ref{app:llm-judge}.

\begin{table}[t]
\centering
\small
\begin{tabular}{p{2.5cm}p{4.5cm}}
\toprule
\textbf{Metric} & \textbf{Definition} \\
\midrule

Description \newline Adherence & Faithfulness to the conditioning description. \\
Meaning & Clarity and semantic depth of the poem. \\
Fluency & Linguistic correctness and naturalness of Arabic. \\
Coherence & Continuity and unity across the poem. \\
Poeticness & Literary quality beyond meter, including imagery and expression. \\
\bottomrule
\end{tabular}
\caption{LLM-based evaluation dimensions.}
\label{tab:eval_metrics}
\end{table}

\paragraph{Human Evaluation.}

We complement the LLM-based evaluation with a blinded human assessment of top-ranked outputs from all four systems. We evaluate Meaning, Fluency, Coherence, and Poeticness using the same 1--5 rubrics as the LLM judges, excluding Description Adherence because it is not applicable to Ashaar and Fanar-Diwan. For each system, we select the five highest-ranked outputs within the shared instance subset based on aggregate multi-LLM scores, yielding 20 poems. Three authors independently rate the selected poems; all are native Arabic speakers, and one additionally has formal training in Arabic prosody and an academic minor in Arabic. Each reported system--dimension score averages 15 ratings across the five poems and three annotators. Further details on selection, blinding, evaluation procedure, and inter-annotator agreement are provided in Appendix~\ref{app:human-evaluation}.

\paragraph{Memorization and Copying Evaluation.}

We additionally assess whether \textsc{Shaer} reproduces content from its training data or from the held-out poems underlying the test instances. Each of the 3,481 generated poems is compared against all 109,070 poems in the SFT training set and against its corresponding held-out source poem after deterministic text normalization.

We measure three complementary indicators of textual overlap: \textit{Exact Match}, which identifies identical normalized poems; \textit{Longest Shared Span}, defined as the longest contiguous sequence of words shared between a generation and a comparison poem; and \textit{5-Gram Jaccard Similarity}, which measures overlap between their sets of word 5-grams. Together, these metrics capture exact reproduction, localized phrase-level reuse, and broader lexical overlap, respectively.

\section{Results and Discussion}
\label{sec:results}

Appendix~\ref{app:qualitative-examples} presents representative poems generated by \textsc{Shaer}, together with their conditioning signals and corresponding meter conformity scores. The remainder of this section focuses on quantitative evaluation.

\subsection{Formal Control Evaluation}

\begin{figure}[t]
    \centering
    \includegraphics[width=\linewidth]{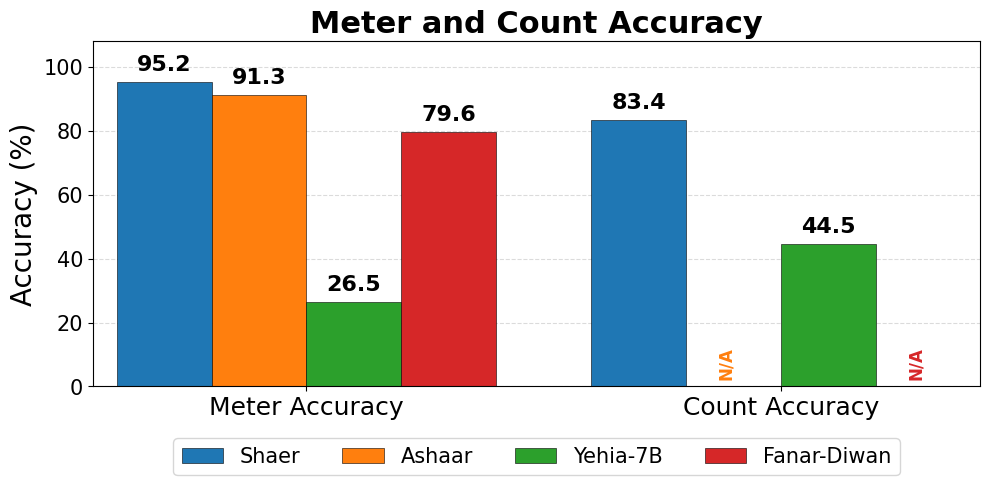}
    \caption{Meter accuracy and count accuracy.}
    \label{fig:meter_count_acc}
\end{figure}

\begin{figure}[t]
    \centering
    \includegraphics[width=\linewidth]{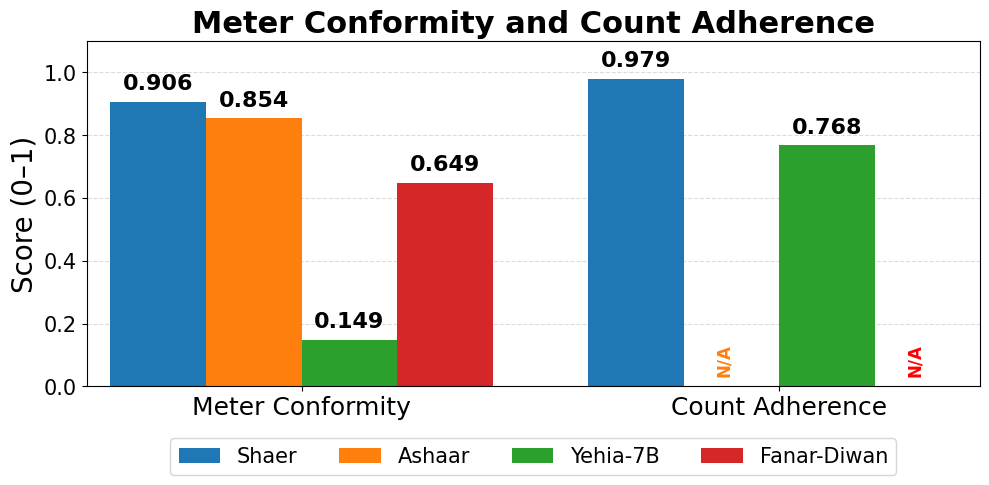}
    \caption{Meter conformity and count adherence scores.}
    \label{fig:meter_conf_count}
\end{figure}

Figures~\ref{fig:meter_count_acc} and~\ref{fig:meter_conf_count} summarize meter and length-control performance.

\textsc{Shaer} achieves the highest base-meter accuracy (95.2\%) and meter-conformity score (0.906), outperforming Ashaar (91.3\%, 0.854), Fanar-Diwan (79.6\%, 0.649), and Yehia-7B (26.5\%, 0.149). It also substantially exceeds Yehia-7B in exact count accuracy (83.4\% vs.\ 44.5\%) and count adherence (0.979 vs.\ 0.768). Because Yehia-7B shares the same foundation model, these differences demonstrate the stronger meter and length control achieved by the jointly conditioned and fine-tuned \textsc{Shaer} system.

Appendix~\ref{app:performance-breakdowns} reports results by base meter and requested-length bucket. \textsc{Shaer} achieves the highest accuracy for 12 of the 13 meters and maintains 93.0--97.0\% base-meter accuracy across length buckets. Its exact count accuracy decreases from 98.4\% for poems of 1--3 verses to 29.0\% for poems of 11--20 verses, but remains substantially above Yehia-7B in every bucket, indicating that longer-form length control remains challenging.

\begin{figure}[t]
    \centering
    \includegraphics[width=\linewidth]{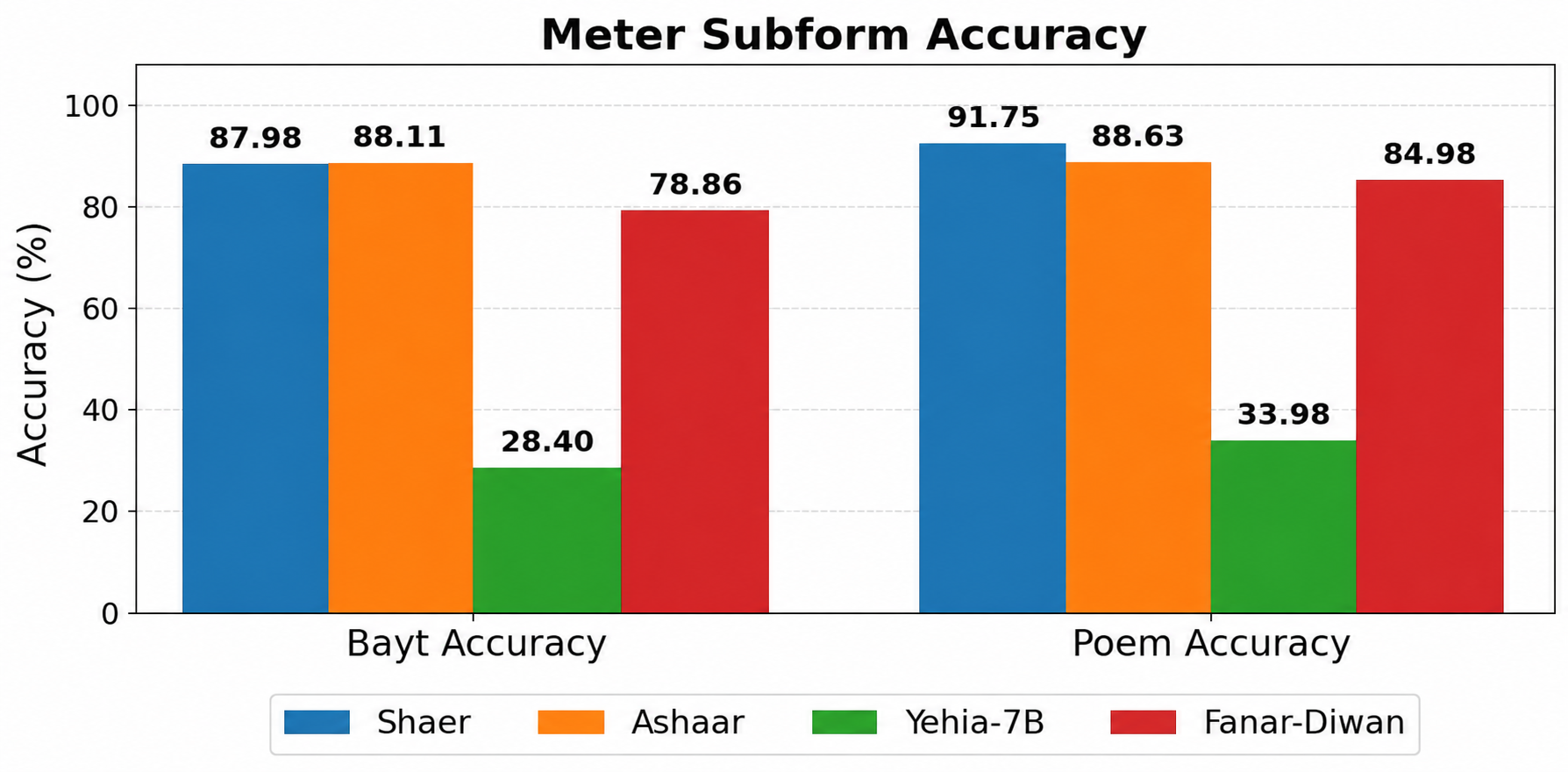}
    \caption{Bayt-level and poem-level meter-subform accuracy.}
    \label{fig:meter_subform_acc}
\end{figure}

Figure~\ref{fig:meter_subform_acc} evaluates fine-grained meter-subform control. Among models explicitly conditioned on a target subform, \textsc{Shaer} achieves a bayt-level accuracy of 87.98\% and a poem-level accuracy of 91.75\%, substantially outperforming Yehia-7B, which obtains bayt-level and poem-level accuracies of 28.40\% and 33.98\%, respectively. Since both systems share the same foundation model and receive the same target subform, this comparison demonstrates the stronger fine-grained prosodic control achieved by the jointly conditioned and fine-tuned \textsc{Shaer} system.

Since Ashaar and Fanar-Diwan do not support subform conditioning, their scores instead measure agreement with the corresponding source poem's subform. Ashaar achieves 88.11\% bayt-level and 88.63\% poem-level agreement, while Fanar-Diwan achieves 78.86\% and 84.98\%, respectively. \textsc{Shaer} attains the highest poem-level accuracy overall, while its bayt-level accuracy is comparable to Ashaar's reference-subform agreement. These results should be interpreted according to the distinct conditioning settings. Additionally, rare subforms are more frequently classified as the dominant Complete (\textarabic{تام}) form; accordingly, evaluation is restricted to target classes represented in the evaluator's 25-class inventory.

\subsection{Semantic and Literary Quality}

\begin{figure*}[t]
    \centering
    \includegraphics[width=\textwidth]{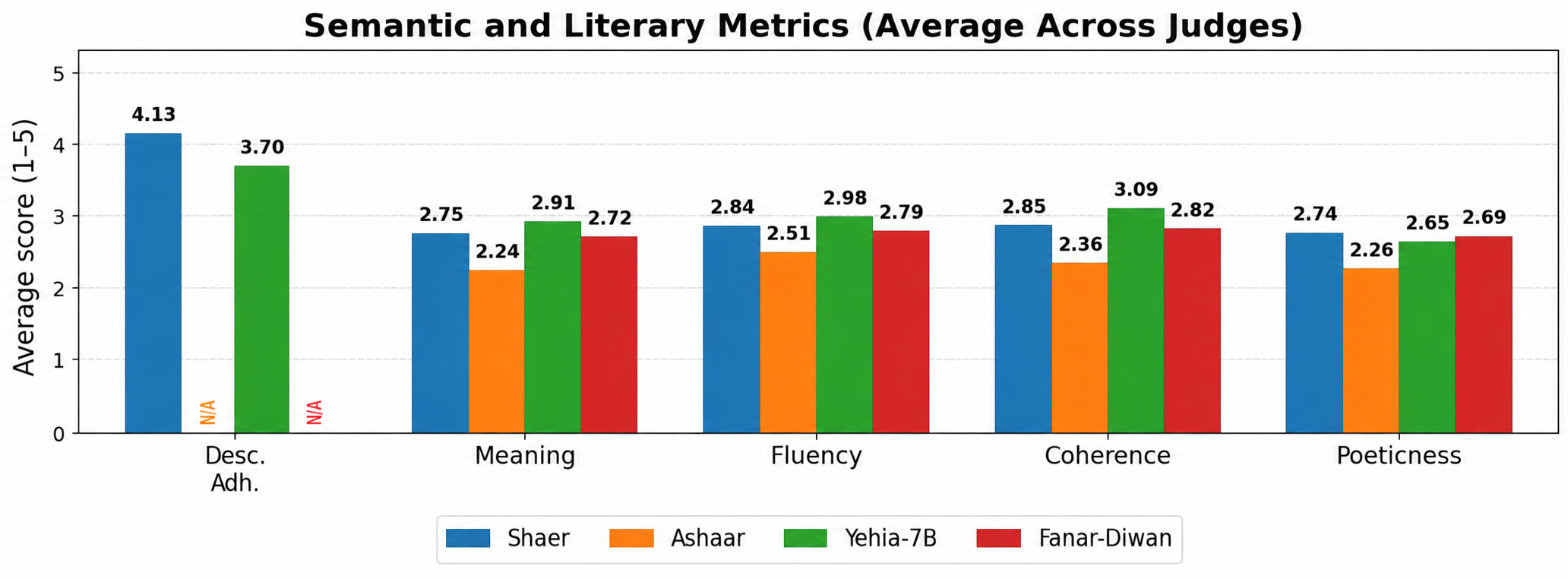}
\caption{Equal-weight mean semantic and literary scores across three LLM judges. Description adherence applies only to description-conditioned systems.}
    \label{fig:multi_judge_metric_summary}
\end{figure*}

Figure~\ref{fig:multi_judge_metric_summary} summarizes semantic and literary quality across the three LLM judges, with judge-specific results reported in Appendix~\ref{app:llm-judge}. Across the four dimensions shared by all systems, \textsc{Shaer} consistently exceeds the two poetry-specific baselines, Ashaar and Fanar-Diwan. The differences relative to Fanar-Diwan are small but consistent across Meaning (2.75 vs.\ 2.72), Fluency (2.84 vs.\ 2.79), Coherence (2.85 vs.\ 2.82), and Poeticness (2.74 vs.\ 2.69), while Ashaar scores lower across all four dimensions. Since these systems use different native conditioning interfaces, these results represent system-level comparisons rather than controlled comparisons of individual modeling choices.

Compared with its unfine-tuned foundation model, Yehia-7B, \textsc{Shaer} achieves higher cross-judge means for Description Adherence (4.13 vs.\ 3.70) and Poeticness (2.74 vs.\ 2.65), whereas Yehia-7B scores higher in Meaning, Fluency, and Coherence. Judge-specific results show the same direction for Description Adherence, Meaning, Coherence, and Poeticness, while Fluency is judge-dependent. Together with the structural results, these findings indicate that task-specific adaptation substantially improves controllability and description adherence while maintaining competitive semantic and literary quality.

\paragraph{Human Evaluation.}

Table~\ref{tab:human-evaluation} reports the complementary human evaluation results. \textsc{Shaer} receives the highest ratings for Fluency (4.20), Coherence (4.20), and Poeticness (4.30), while Yehia-7B receives the highest Meaning rating (4.13, compared with 4.00 for \textsc{Shaer}). Fanar-Diwan receives intermediate ratings, whereas Ashaar receives the lowest mean ratings across all four dimensions.

The human assessment partially aligns with the multi-LLM evaluation: both favor \textsc{Shaer} over Yehia-7B on Poeticness and Yehia-7B over \textsc{Shaer} on Meaning, while the human annotators additionally favor \textsc{Shaer} on Fluency and Coherence. Given that this evaluation targets top-ranked outputs rather than a representative test-set sample, these findings provide complementary evidence about the systems' strongest identified generations rather than their average performance.

\begin{table}[t]
\centering
\small
\setlength{\tabcolsep}{4pt}
\begin{tabular}{lcccc}
\toprule
\textbf{System} &
\textbf{Mean.} &
\textbf{Flu.} &
\textbf{Coh.} &
\textbf{Poet.} \\
\midrule
\textsc{Shaer} & 4.00 & \textbf{4.20} & \textbf{4.20} & \textbf{4.30} \\
Yehia-7B       & \textbf{4.13} & 3.50 & 3.80 & 3.30 \\
Fanar-Diwan    & 3.80 & 3.70 & 3.80 & 3.50 \\
Ashaar         & 3.10 & 3.00 & 3.60 & 2.40 \\
\bottomrule
\end{tabular}
\caption{Blinded human evaluation of five top-ranked poems per system. Scores are means over 15 ratings (five poems $\times$ three annotators) on a 1--5 scale.}
\label{tab:human-evaluation}
\end{table}

\subsection{Memorization and Copying Analysis}

No exact normalized copies were found among the 3,481 \textsc{Shaer} generations. One generation (0.0287\%) shared at least 13 consecutive words with a training poem, and four (0.1149\%) with their paired held-out sources; none shared 30 or more words in either comparison. Median word 5-gram Jaccard similarity was 0.0 for both comparisons, with maxima of 0.119 and 0.220, respectively. Manual inspection revealed localized reuse or paraphrasing rather than full-poem reproduction. We therefore find no evidence of systematic memorization or wholesale copying, although rare source-associated phrase reuse occurs.

\section{Conclusion}

We presented \textsc{Shaer}, a controllable Classical Arabic poetry generation framework jointly conditioned on natural-language descriptions, meter subforms, and target hemistich counts. To support this task, we constructed an enriched corpus of 116,032 classical Arabic poems derived from Ashaar, containing normalized meter-subform labels and automatically generated, validated semantic descriptions. Building on this resource, we adapted Yehia-7B through QLoRA-based supervised fine-tuning with a completion-only objective.

Across the 3,481 held-out test instances, \textsc{Shaer} achieves 95.17\% base-meter accuracy, 91.75\% poem-level meter-subform accuracy, and 83.40\% exact count accuracy, representing substantial gains over its untuned foundation model. It also attains the highest base-meter accuracy among all evaluated systems. Evaluation with three LLM judges indicates that \textsc{Shaer} exceeds the poetry-specific comparison systems across the shared semantic and literary dimensions, while its comparison with Yehia-7B reveals a trade-off between stronger controllability and performance on some unconstrained literary-quality dimensions. A blinded human assessment of top-ranked outputs further supports the model's competitive literary quality. In addition, memorization analysis finds no exact normalized copies from either the training corpus or the paired held-out source poems.

Together, these findings demonstrate that \textsc{Shaer} provides reliable joint control over semantic intent, fine-grained prosodic form, and poem length while preserving competitive semantic and literary quality. The released code, models, annotations, and evaluation resources provide a foundation for further research on controllable Arabic literary generation.

Future work includes explicit rhyme conditioning and evaluation, component-wise ablations, broader human evaluation on representative samples, improved handling of rare meter subforms and longer poems, and extension to additional Arabic poetic genres.

\section{Limitations}

Despite the strong results achieved by \textsc{Shaer}, several limitations remain. First, this work focuses on Classical Arabic poetry written in the 13 canonical meters retained after corpus filtering. Consequently, the dataset and model do not cover free verse, prose poetry, colloquial poetry, or other contemporary poetic forms. Performance is also less certain for rare meter subforms and longer requested poems because of their limited representation.

Second, 95.6\% of the corpus lacked explicit meter-subform metadata and was assigned the canonical Complete (\textarabic{تام}) form by default. Although manual inspection and classifier-based validation support this convention, these labels remain partly heuristic.

Third, semantic descriptions are generated automatically rather than written by human annotators. Despite structural validation, corpus-wide quality checks, and human evaluation of sampled descriptions, some residual semantic noise may remain.

Fourth, the evaluation relies partly on learned meter classifiers and LLM judges. Although these evaluators were validated and three LLM judges were used, their scores remain approximations of expert human judgment. The complementary human evaluation is also limited to selected top-ranked outputs rather than a representative test-set sample.

Finally, rhyme is learned implicitly but is neither explicitly controlled nor quantitatively evaluated. Because the experiments do not include component-wise ablations, the observed improvements should be attributed to the jointly conditioned and fine-tuned \textsc{Shaer} system rather than to any individual control mechanism.

\bibliography{custom_acl_literal_et_al}

\appendix

\vspace{1em}
\noindent\textbf{\large Appendix}

\section{Background on Classical Arabic Poetry}
\label{app:poetry-background}

\subsection{Poetic Structure}

Classical Arabic poetry follows a highly structured form governed by meter and
rhyme. A poem (\textarabic{قصيدة}, \emph{qasida}) consists of verses
(\textarabic{أبيات}, \emph{abyat}; singular \textarabic{بيت}, \emph{bayt}),
each typically divided into two hemistichs (\textarabic{شطرين},
\emph{shatrayn}): \textarabic{الصدر} (\emph{sadr}, first hemistich) and
\textarabic{العجز} (\emph{ajuz}, second hemistich). Hemistichs within a poem
typically conform to a single meter (\textarabic{بحر}, \emph{bahr}), while
the verses share a common rhyme (\textarabic{قافية}, \emph{qafiyah}),
realized at the end of the second hemistich \cite{abbas2001metrical}. These
constraints require poetry generation to maintain both semantic and prosodic
consistency.

\subsection{Meter and Meter Subforms}

Arabic prosody (\textarabic{علم العروض}, \emph{ʿilm al-ʿarud}) describes the
metrical organization of classical Arabic poetry. Each meter
(\textarabic{بحر}, \emph{bahr}) is characterized by an ordered sequence of
recurring metrical feet (\textarabic{تفاعيل}, \emph{tafaʿil})
\cite{abbas2001metrical}. For example, the canonical pattern of Rajaz
(\textarabic{الرجز}) is
\textarabic{مستفعلن مستفعلن مستفعلن} in each hemistich.
Table~\ref{tab:meter-keys} summarizes the sixteen canonical Khalilian meters
and their meter keys.

\begin{table*}[t]
\centering
\small
\begin{tabular}{lll}
\hline
\textbf{English Name} & \textbf{Arabic Name} & \textbf{Meter Key} \\
\hline
Taweel     & \textarabic{الطويل}     & \textarabic{فعولن مفاعيلن فعولن مفاعيلن} \\
Madeed     & \textarabic{المديد}     & \textarabic{فاعلاتن فاعلن فاعلاتن} \\
Baseet     & \textarabic{البسيط}     & \textarabic{مستفعلن فاعلن مستفعلن فاعلن} \\
Wafer      & \textarabic{الوافر}     & \textarabic{مفاعلتن مفاعلتن فعولن} \\
Kamel      & \textarabic{الكامل}     & \textarabic{متفاعلن متفاعلن متفاعلن} \\
Hazaj      & \textarabic{الهزج}      & \textarabic{مفاعيلن مفاعيلن} \\
Ramal      & \textarabic{الرمل}      & \textarabic{فاعلاتن فاعلاتن فاعلاتن} \\
Rajaz      & \textarabic{الرجز}      & \textarabic{مستفعلن مستفعلن مستفعلن} \\
Sare'e     & \textarabic{السريع}     & \textarabic{مستفعلن مستفعلن فاعلن} \\
Munsareh   & \textarabic{المنسرح}    & \textarabic{مستفعلن مفعولاتُ مستفعلن} \\
Khafeef    & \textarabic{الخفيف}     & \textarabic{فاعلاتن مستفعلن فاعلاتن} \\
Mujtath    & \textarabic{المجتث}     & \textarabic{مستفعلن فاعلاتن} \\
Mutakareb  & \textarabic{المتقارب}   & \textarabic{فعولن فعولن فعولن فعولن} \\
Mutadarak  & \textarabic{المتدارك}   & \textarabic{فاعلن فاعلن فاعلن فاعلن} \\
Muqtadheb  & \textarabic{المقتضب}    & \textarabic{فاعلات مفتعلن} \\
Mudhare'e  & \textarabic{المضارع}    & \textarabic{مفاعيلن فاعلاتن} \\
\hline
\end{tabular}
\caption{The sixteen canonical Khalilian meters and their corresponding
meter keys \cite{alabbas-etal-2014-basrah}.}
\label{tab:meter-keys}
\end{table*}

A base meter can admit multiple valid realizations through systematic
modifications of its canonical structure \cite{atiq-arud-qafiyah}. We refer
to these finer-grained realizations as \emph{meter subforms}. Common examples
include Complete (\textarabic{تام}), Catalectic (\textarabic{مجزوء}),
Hemistich (\textarabic{مشطور}), and Severely Truncated
(\textarabic{منهوك}); these are illustrative rather than exhaustive, and
valid subforms vary across meters. Consequently, two poems can share the same
base meter while following different subforms. Table~\ref{tab:rajaz-subforms}
illustrates this distinction using Rajaz.

\begin{table*}[t]
\centering
\footnotesize
\setlength{\tabcolsep}{3.5pt}
\renewcommand{\arraystretch}{1.25}

\begin{tabularx}{\textwidth}{
    @{}
    >{\centering\arraybackslash}p{2.6cm}
    >{\centering\arraybackslash}p{0.8cm}
    >{\centering\arraybackslash}p{4.2cm}
    X
    >{\centering\arraybackslash}p{4.2cm}
    @{}
}
\hline

\textbf{Name} &
\makecell{\textbf{\#}\\\textbf{Feet}} &
\textbf{Metrical Feet Pattern} &
\multicolumn{1}{c}{\textbf{Definition}} &
\textbf{Example} \\

\hline

\makecell[c]{
Complete\\
\textarabic{تام}
}
&
6
&
\makecell[c]{
\textarabic{مُسْتَفْعِلُنْ مُسْتَفْعِلُنْ مُسْتَفْعِلُنْ}\\
\textarabic{مُسْتَفْعِلُنْ مُسْتَفْعِلُنْ مُسْتَفْعِلُنْ}
}
&
Retains the full three-foot pattern in each of the two hemistichs.
&
\makecell[r]{
\textarabic{لَا خَيْرَ فِيمَنْ كَفَّ عَنَّا شَرَّهُ}\\
\textarabic{إِنْ كَانَ لَا يُرْجَى لِيَوْمٍ خَيْرُهُ}
}
\\[0.7em]

\makecell[c]{
Truncated\\
\textarabic{مجزوء}
}
&
4
&
\makecell[c]{
\textarabic{مُسْتَفْعِلُنْ مُسْتَفْعِلُنْ}\\
\textarabic{مُسْتَفْعِلُنْ مُسْتَفْعِلُنْ}
}
&
Retains two feet in each of the two hemistichs.
&
\makecell[r]{
\textarabic{فَارفِقْ بِصَبٍّ ذِي جَوى}\\
\textarabic{مِنْ حَرِّهِ ذَابَ الحَشَى}
}
\\[0.7em]

\makecell[c]{
Single-Hemistich\\
\textarabic{مشطور}
}
&
3
&
\makecell[c]{
\textarabic{مُسْتَفْعِلُنْ مُسْتَفْعِلُنْ مُسْتَفْعِلُنْ}
}
&
Consists of a single three-foot verse.
&
\makecell[r]{
\textarabic{إِنَّ ابْنَتِي بَيْضاءُ مِنْ بِيضٍ زُهُرْ}
}
\\[0.7em]

\makecell[c]{
Exhausted\\
\textarabic{منهوك}
}
&
2
&
\makecell[c]{
\textarabic{مُسْتَفْعِلُنْ مُسْتَفْعِلُنْ}
}
&
Consists of a single two-foot verse.
&
\makecell[r]{
\textarabic{يَا لَيْتَنِي فِيهَا جَذَعْ}
}
\\

\hline
\end{tabularx}

\caption{Illustrative meter subforms of Rajaz
(\textarabic{بحر الرجز}), showing the number of metrical feet per verse,
their metrical patterns, and corresponding examples. The set is not exhaustive.}

\label{tab:rajaz-subforms}
\end{table*}

\section{Analysis of the Original Ashaar Corpus}
\label{app:ashaar-analysis}

The original Ashaar corpus contains 254,630 poems and 12 metadata fields per poem. However, substantial inconsistencies in metadata coverage, meter annotations, description availability, poem length, and textual quality motivated the cleaning and filtering procedures described in Section~\ref{sec:data}. Table~\ref{tab:ashaar-analysis} summarizes the key statistics.

\begin{table*}[t]
\centering
\small
\begin{tabular}{p{0.68\textwidth}r}
\toprule
\textbf{Statistic} & \textbf{Value} \\
\midrule
Total number of poems & 254,630 \\
Number of metadata fields & 12 \\
Number of unique meter labels & 101 \\
Poems lacking meter annotations & 39.77\% \\
Poems containing at least one empty hemistich & 0.17\% \\
Number of unique theme categories & 19 \\
Poems lacking theme annotations & 73.48\% \\
Poems with unreliable language-type labels (\textarabic{عامي}, \textarabic{شعبي}, missing, or ``-'') & 31.44\% \\
Poems lacking descriptions & 94.32\% \\
Average number of hemistichs per poem & 30.30 \\
Median number of hemistichs per poem & 14 \\
Minimum hemistich count & 0 \\
Maximum hemistich count & 11,608 \\
Poems exceeding 40 hemistichs (20 verses) & 21.68\% \\
Poems containing Latin characters, numerical digits, or URLs & 2.56\% \\
Poems containing \textit{tatweel} characters & 45.45\% \\
\bottomrule
\end{tabular}
\caption{Key characteristics of the original Ashaar corpus before preprocessing and filtering.}
\label{tab:ashaar-analysis}
\end{table*}

\section{Data Cleaning and Preprocessing}
\label{app:data-cleaning}

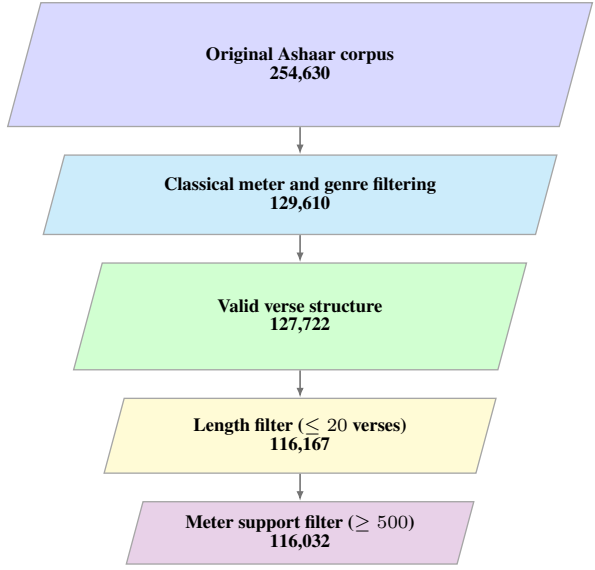
\begin{figure}[h] \centering \scriptsize \begin{tikzpicture}
[ node distance=0.36cm, 
  stage/.style={ trapezium, trapezium left angle=75, trapezium right angle=105, draw=black!35, line width=0.5pt, align=center, minimum height=0.4cm, inner sep=2.5pt, font=\bfseries },
  arrow/.style={ -{Latex[length=1.3mm]}, semithick, draw=black!55 } ] 

   \node[stage, fill=blue!15, minimum width=7.8cm] (s1) {Original Ashaar corpus\\[-1]254,630}; 
   \node[stage, fill=cyan!18, minimum width=6.8cm, below=of s1] (s2) {Classical meter and genre filtering\\[-1]129,610};
   \node[stage, fill=green!18, minimum width=6cm, below=of s2] (s3) {Valid verse structure\\[-1]127,722};
   \node[stage, fill=yellow!20, minimum width=5.2cm, below=of s3] (s4) {Length filter ($\leq 20$ verses)\\[-1]116,167}; 
   \node[stage, fill=violet!18, minimum width=4.6cm, below=of s4] (s5) {Meter support filter ($\geq 500$)\\[-1]116,032}; 
   
   \draw[arrow] (s1) -- (s2);
   \draw[arrow] (s2) -- (s3); 
   \draw[arrow] (s3) -- (s4);   
   \draw[arrow] (s4) -- (s5);
\end{tikzpicture}
\caption{Corpus filtering pipeline. Starting from the original Ashaar corpus, the filtering process yields a final dataset of 116,032 classical Arabic poems.}
\label{fig:filtering-pipeline}
\end{figure}

We apply a four-stage pipeline to remove structural, orthographic, and metadata inconsistencies from the original Ashaar corpus. First, the corpus is restricted to classical Arabic poetry written in the sixteen canonical Khalilian meters. Poems labeled as colloquial or non-classical (\textarabic{شعبي}, \textarabic{عامي}), as well as poems lacking meter annotations, are discarded. Second, poems with malformed verse structures, empty hemistichs, or hemistichs containing Latin characters, digits, or URLs are removed. Internal elongation (\textarabic{تطويل (-)}) characters are also stripped, while edge elongation and all diacritics are preserved. Third, poems exceeding twenty verses are excluded to align the corpus with the supervised fine-tuning maximum sequence length of 1,024 tokens. A post-tokenization audit using the final instruction format found that only 4 of the 109,070 training sequences (0.0037\%) still exceeded this limit, by 5--25 tokens, while no validation or test sequence exceeded it. Finally, meter annotations are normalized to the canonical meter inventory, and base meters represented by fewer than 500 poems are removed. Figure~\ref{fig:filtering-pipeline} summarizes the resulting counts at each stage.

\section{Meter Subform Details}
\label{app:meter-subform-details}

\subsection{Default-Label Validation}
\label{app:subform-validation}

We first manually assessed the default Complete assignment on a sample of 1,000 poems. An author with formal training in Arabic prosody inspected the sampled poems and confirmed the Complete form in 97.2\% of cases. This manual check evaluates the plausibility of the defaulting convention on a sample; it does not constitute exhaustive re-annotation of the 110,926 default-labeled poems.

We reproduce the meter-subform classification setting of AraPoemBERT \cite{qarah-2024-arapoembert} using the dataset released by the authors, as the publicly released checkpoint could not be directly loaded as a complete sequence-classification model. Our reproduction achieved 97.38\% accuracy, comparable to the 97.79\% reported in the original work. We apply the resulting classifier to all poems assigned the default Complete label. Table~\ref{tab:complete-validation-by-meter} reports the resulting agreement by base meter.

The four meters with low agreement reflect limitations of the classifier itself. Complete forms of Madeed, Hazaj, and Mujtath are absent from the classifier's released class inventory, while Rajaz Complete has a reported precision of 0.4725 \cite{qarah-2024-arapoembert}. In contrast, agreement ranges from 98.76\% to 99.93\% across the remaining nine meters. We use these predictions only to assess the default-labeling strategy and do not use them to modify the corpus annotations.

\begin{table*}[t]
\centering
\small
\begin{tabular}{lrrrr}
\toprule
\textbf{Base Meter} &
\textbf{Default Complete} &
\textbf{Predicted Complete} &
\textbf{Predicted Non-Complete} &
\textbf{Agreement} \\
\midrule
Taweel    & 28,028 & 28,007 & 21    & 99.93\% \\
Mutakareb & 5,189  & 5,185  & 4     & 99.92\% \\
Baseet    & 16,811 & 16,787 & 24    & 99.86\% \\
Kamel     & 19,438 & 19,399 & 39    & 99.80\% \\
Khafeef   & 9,555  & 9,531  & 24    & 99.75\% \\
Wafer     & 11,097 & 11,047 & 50    & 99.55\% \\
Sare'e    & 7,190  & 7,154  & 36    & 99.50\% \\
Ramal     & 3,643  & 3,615  & 28    & 99.23\% \\
Munsareh  & 2,747  & 2,713  & 34    & 98.76\% \\
Madeed    & 538    & 37     & 501   & 6.88\% \\
Rajaz     & 3,973  & 225    & 3,748 & 5.66\% \\
Hazaj     & 609    & 15     & 594   & 2.46\% \\
Mujtath   & 2,060  & 33     & 2,027 & 1.60\% \\
\midrule
\textbf{Overall} &
\textbf{110,926} &
\textbf{103,748} &
\textbf{7,178} &
\textbf{93.53\%} \\
\bottomrule
\end{tabular}

\caption{Agreement between the default Complete annotations and predictions of the reproduced AraPoemBERT meter-subform classifier, reported by base meter.}
\label{tab:complete-validation-by-meter}
\end{table*}

\subsection{Meter--Subform Distribution}
\label{app:subform-distribution}

Table~\ref{tab:meter-subform-distribution} reports the distribution of the 21 meter--subform combinations in the final corpus.

\begin{table}[t]
\centering
\small
\begin{tabular}{llr}
\toprule
\textbf{Base Meter} & \textbf{Subform} & \textbf{Poems} \\
\midrule
Baseet    & Complete  & 16,811 \\
Baseet    & Mukhalla' & 651 \\
Baseet    & Majzu'    & 159 \\
Khafeef   & Complete  & 9,555 \\
Khafeef   & Majzu'    & 298 \\
Rajaz     & Complete  & 3,973 \\
Rajaz     & Majzu'    & 612 \\
Ramal     & Complete  & 3,643 \\
Ramal     & Majzu'    & 949 \\
Sare'e    & Complete  & 7,190 \\
Taweel    & Complete  & 28,028 \\
Kamel     & Complete  & 19,438 \\
Kamel     & Majzu'    & 1,862 \\
Kamel     & Ahadh     & 373 \\
Mutakareb & Complete  & 5,189 \\
Mujtath   & Complete  & 2,060 \\
Madeed    & Complete  & 538 \\
Munsareh  & Complete  & 2,747 \\
Hazaj     & Complete  & 609 \\
Wafer     & Complete  & 11,097 \\
Wafer     & Majzu'    & 250 \\
\bottomrule
\end{tabular}

\caption{Distribution of meter--subform combinations in the final corpus.}
\label{tab:meter-subform-distribution}
\end{table}

\section{Semantic Description Generation and Validation}
\label{app:description-generation}

The description-generation pipeline consists of three components: selection of the generation configuration, specification of the prompting and decoding procedure for the selected configuration, and validation of the resulting corpus descriptions.

\subsection{Generation Configuration Selection}
\label{app:description-configuration-selection}

Before generating descriptions for the full corpus, we compare three candidate generation configurations differing in generative model and prompting strategy. For each configuration, we generate descriptions for a stratified sample of 600 poems covering diverse base meters, meter subforms, and poem lengths.

Each generated description is independently assigned a Pass or Fail label by three of the authors, who are blinded to the configuration that produced it. All three annotators are native Arabic speakers; one additionally has formal training in Arabic prosody. Descriptions are assessed according to nine criteria: fluency, faithfulness to the poem content, coverage of the central meaning and imagery, absence of hallucinated content, absence of direct verse copying, sufficient specificity, use of the required opening phrase (\textarabic{القصيدة تتحدث عن...}), single-paragraph formatting, and absence of metadata leakage. A description receives a Fail label if any criterion is violated.

Disagreements are resolved by majority vote. Inter-annotator agreement across the evaluations is Fleiss' $\kappa = 0.817$, indicating strong consistency in the application of the evaluation rubric.

The three candidate configurations achieve pass rates of 428/600 (71.3\%), 506/600 (84.3\%), and 557/600 (92.8\%), respectively. We select the third configuration, which achieves the highest pass rate, for corpus-wide description generation.

\subsection{Prompting and Decoding Configuration}
\label{app:description-prompting}

The selected configuration uses Qwen3.5-35B-A3B \cite{qwen-2026-qwen35} under deterministic decoding. The system prompt instructs the model to produce a single descriptive Arabic paragraph beginning with \textarabic{(القصيدة تتحدث عن...)}, capturing the poem's central meaning, imagery, and tone. It further instructs the model to avoid copying verses, introducing unsupported content, or referring to formal properties such as meter, rhyme, or poem length. The response is returned as a JSON object containing a single \texttt{new\_description} field.

The user prompt provides all hemistichs of the source poem followed by a one-line instruction requesting the semantic description.

\begin{quote}
\footnotesize
\begin{Arabic}
الأبيات:
\end{Arabic}

\medskip
\begin{flushright}
\texttt{[All poem hemistichs]}
\end{flushright}

\medskip
\begin{Arabic}
اكتب وصفًا واحدًا يلتقط أفكار القصيدة ومعناها وصورها ونبرتها العامة.
\end{Arabic}
\end{quote}

The decoding configuration used for corpus-wide generation is reported in Table~\ref{tab:desc-gen-config}.

\begin{table}[t]
\centering
\small
\begin{tabular}{ll}
\toprule
\textbf{Parameter} & \textbf{Value} \\
\midrule
Model & \texttt{Qwen/Qwen3.5-35B-A3B} \\
Temperature & 0.0 \\
Maximum output tokens & 300 \\
Response format & JSON object \\
Reasoning effort & Disabled \\
Provider fallbacks & Disabled \\
\bottomrule
\end{tabular}
\caption{Decoding configuration used for corpus-wide semantic description generation.}
\label{tab:desc-gen-config}
\end{table}

\subsection{Description Validation}
\label{app:description-validation}

Following configuration selection, descriptions generated for the full corpus undergo a two-stage validation procedure. The first stage ensures structural validity during generation, while the second performs fine-grained poem-level quality checks over the generated descriptions. Descriptions that fail verification are regenerated and re-evaluated before inclusion in the final corpus.

\subsubsection{Stage 1: Automatic Structural Validation}

Each description is validated immediately after generation. The pipeline parses the JSON response, verifies the presence of the \texttt{new\_description} field, normalizes whitespace, and rejects empty, malformed, or schema-leaking outputs. Failed generations are retried up to four times. Publication is blocked if any poem is missing a valid description. This stage ensures structural correctness and does not assess semantic quality.

\subsubsection{Stage 2: Corpus-Wide Fine-Grained Validation}

We subsequently perform deterministic poem-level quality checks over the complete set of generated poem-description pairs. These checks assess adherence to the required output format, leakage of formal poem metadata, and direct copying from the source poem. Table~\ref{tab:combined-verification} reports first-pass results before regeneration. Descriptions failing one or more checks were regenerated and re-evaluated, and all descriptions retained in the final corpus passed the Stage~2 checks.

Format-compliance checks verify that descriptions are non-empty, begin with the required opening phrase, consist of a single paragraph, are not formatted as lists, do not take the form of questions, requests, or commands, and contain no leaked JSON syntax. Metadata-leakage checks identify references to formal poetic properties that descriptions were instructed not to mention. Copying checks detect full hemistich reproduction, long copied phrases, and excessive overlap with the source poem.

\begin{table*}[t]
\centering
\small
\begin{tabular}{llc}
\toprule
\textbf{Category} & \textbf{Metric} & \textbf{Pass Rate} \\
\midrule
\multirow{6}{*}{Format Compliance}
  & Non-empty description          & 100\%   \\
  & Required opening phrase        & 100\%   \\
  & Single paragraph               & 100\%   \\
  & Not a list                     & 100\%   \\
  & Not a question/request/command & 99.29\% \\
  & No JSON leakage                & 99.99\% \\
\midrule
Metadata Leakage
  & No metadata leakage            & 98.34\% \\
\midrule
\multirow{3}{*}{Copying Detection}
  & No full hemistich copied       & 97.46\% \\
  & No long phrase copied          & 97.43\% \\
  & No excessive copying           & 99.49\% \\
\bottomrule
\end{tabular}
\caption{First-pass corpus-level validation results for automatically generated semantic descriptions, reported before regeneration. Pass rates measure compliance with formatting requirements, absence of metadata leakage, and avoidance of direct copying from source poems. Failed descriptions were regenerated and re-evaluated before inclusion in the final corpus.}
\label{tab:combined-verification}
\end{table*}

\section{Final Dataset Characteristics}
\label{app:final-dataset}

The released artifacts distinguish between newly created resources and inherited corpus content. Our code, adapters, and separable annotations and semantic descriptions are released under the Apache-2.0 license. The underlying poem text remains subject to Ashaar's non-commercial and research-use restrictions and reserved takedown rights; our release does not relicense that content.

Table~\ref{tab:dataset_stats} summarizes the principal characteristics of the final dataset.

\begin{table}[h]
\centering
\small
\begin{tabular}{lr}
\toprule
\textbf{Statistic} & \textbf{Value} \\
\midrule
Poems                          & 116,032 \\
Base meters                    & 13 \\
Meter--subform pairs            & 21 \\
Average verses per poem        & 5.52 \\
Median verses per poem         & 4 \\
Generated descriptions         & 116,032 \\
Average description length     & 266.2 chars \\
\bottomrule
\end{tabular}
\caption{Summary statistics of the final \textsc{Shaer} dataset after preprocessing, filtering, meter subform annotation, and semantic description generation.}
\label{tab:dataset_stats}
\end{table}

\section{Supervised Fine-Tuning Details}
\label{app:sft-details}

Table~\ref{tab:hyperparams} summarizes the training hyperparameters and runtime settings used during QLoRA fine-tuning, and the instruction templates used during dataset construction are provided below.

\begin{table*}[t]
\centering
\small
\begin{tabular}{p{0.28\textwidth}p{0.66\textwidth}}
\toprule
\textbf{Parameter} & \textbf{Value} \\
\midrule
Base model & \texttt{Navid-AI/Yehia-7B-preview} \\
Training objective & Causal language modeling with completion-only loss \\
Prompt loss masking & Enabled; prompt tokens are masked with \texttt{-100} \\
Max sequence length & 1024 \\
Train packing & Enabled \\
Truncation mode & \texttt{keep\_end} \\
Tokenizer padding side & Left \\
Quantization & QLoRA, 4-bit NF4, double quantization, bfloat16 \\
LoRA target modules & \texttt{all-linear} \\
LoRA rank / alpha / dropout & 64 / 128 / 0.05 \\
LoRA bias & \texttt{none} \\
LoRA task type & \texttt{CAUSAL\_LM} \\
RS-LoRA & Enabled \\
Optimizer & \texttt{paged\_adamw\_8bit} \\
Learning rate & \(8\times10^{-5}\) \\
Scheduler & Cosine \\
Warmup ratio & 0.03 \\
Weight decay & 0.0 \\
Max gradient norm & 1.0 \\
Batch size per device & 1 \\
Gradient accumulation & 16 \\
Effective batch size & 16 \\
Epochs & 2 \\
Global steps & 4,236 \\
Evaluation interval & 200 steps \\
Save interval & 200 steps \\
Best-model metric & \texttt{eval\_loss} \\
Seed & 42 \\
Train/val/test split & 94 / 3 / 3 (stratified by base meter, meter subform, and length bucket) \\
Length buckets & 1 to 3, 4 to 6, 7 to 10, and 11 to 20 verses \\
Training sampler & Weighted sampler with \(w(g) \propto (1/\mathrm{count}(g))^{0.6}\) over (base meter, meter subform, length bucket) \\
Runtime & 12.53 hours \\
\bottomrule
\end{tabular}
\caption{Training hyperparameters, optimization settings, and runtime configuration used for QLoRA-based supervised fine-tuning of \textsc{Shaer}.}
\label{tab:hyperparams}
\end{table*}

\subsection{Instruction Templates}

Training examples are constructed using a fixed instruction format. The system prompt defines the model's role as a classical Arabic poet and enforces strict meter adherence, while the user prompt supplies the target meter, meter subform, hemistich count, and semantic description.

\subsubsection{System Prompt}

\begin{quote}
\begin{Arabic}
أنت شاعر عربي تكتب الشعر العمودي الكلاسيكي.\\
التزم بالبحر المحدد في كل شطر، واستلهم من الموضوع دون نقله حرفياً.\\
أخرج الأبيات فقط دون مقدمة أو تعليق.\\
التزم التزاماً صارماً بالبحر المطلوب، ولا تخرج عنه.
\end{Arabic}
\end{quote}

\subsubsection{User Prompt Template}

\begin{quote}
\small
\begin{flushright}
\begin{tabular}{@{}r l@{}}
\texttt{\{base\_meter\}} & \textarabic{البحر الأساسي:} \\
\texttt{\{form\}} & \textarabic{الصيغة:} \\
\texttt{\{num\_lines\}} & \textarabic{عدد الأشطر المطلوب:} \\
\texttt{\{description\}} & \textarabic{الموضوع:} \\
\end{tabular}
\end{flushright}

\vspace{4pt}
\begin{Arabic}
اكتب
\texttt{\{senil\_mun\}}
أشطارًا ملتزمة بصيغة
\texttt{\{mrof\}}
من بحر
\texttt{\{retem\_esab\}}
دون أي شرح إضافي.
\end{Arabic}
\end{quote}

During training, placeholders are replaced with the corresponding poem metadata and semantic description. The loss is computed only over the completion tokens.

\section{Native Input Representation}
\label{app:native_input_representation}

Table~\ref{tab:native_inputs} summarizes the native input representations and conditioning signals used by the evaluated models.

\begin{table*}[t]
\centering
\footnotesize
\setlength{\tabcolsep}{3pt}
\begin{tabular}{
p{0.13\textwidth}
p{0.34\textwidth}
p{0.28\textwidth}
p{0.15\textwidth}
}
\toprule
\textbf{Model} &
\textbf{Native Input Format} &
\textbf{Conditioning Signals} &
\textbf{Output} \\
\midrule

Yehia-7B \cite{yehia2025} &
Instruction-style prompt using the same format as \textsc{Shaer}, but without task-specific fine-tuning. &
Meter, meter subform, hemistich count, and semantic description provided through prompting only. &
Generated poem \\
\midrule

Ashaar \cite{alyafeai-etal-2023-ashaar} &
Structured HTML-like representation using special tokens for meter, theme, qafiyah, poem separators, bayt separators, and verse separators. &
Base meter, theme, and rhyme/qafiyah. &
\makecell[l]{Generated or\\completed poem} \\
\midrule

Fanar-Diwan \cite{fanar2026fanar20} &
Metadata-conditioned prompt consisting of a hemistich followed by bracketed poetic metadata, such as meter, topic, era, poet identifier, and rhyme letter. &
Base meter, topic, era, poet identity, and rhyme letter. &
\makecell[l]{Poem\\continuation} \\
\midrule

\textsc{Shaer} &
Arabic instruction-following prompt explicitly specifying the target meter, meter subform, number of hemistichs, and semantic description. &
Meter, meter subform, hemistich count, and semantic description. &
\makecell[l]{Controlled\\generated\\poem} \\

\bottomrule
\end{tabular}
\caption{Native input representations, conditioning signals, and output formats of the evaluated models.}
\label{tab:native_inputs}
\end{table*}

\section{Poem Generation Decoding}
\label{app:generation-decoding}

Table~\ref{tab:poem-generation-decoding} reports the decoding and stopping settings used to generate the evaluated poems. 

\begin{table*}[t]
\centering
\footnotesize
\setlength{\tabcolsep}{4pt}
\begin{tabular}{lcccccp{5.0cm}}
\toprule
\textbf{System} &
\textbf{Temperature} &
\textbf{Top-$p$} &
\textbf{Top-$k$} &
\shortstack{\textbf{Maximum new}\\\textbf{tokens}} &
\shortstack{\textbf{Repetition}\\\textbf{penalty}} &
\textbf{Stopping and generation-time post-processing} \\
\midrule

\textsc{Shaer}
& 0.55
& 0.90
& Not set
& 768
& 1.08
& Model EOS or the 768-token generation limit; no custom stopping strings or token IDs. \\

Yehia-7B
& 0.55
& 0.90
& Not set
& 768
& 1.08
& Model EOS or the 768-token generation limit; no custom stopping strings or token IDs. \\

Fanar-Diwan
& 0.55
& 0.90
& Not set
& 768
& 1.08
& Model EOS or the 768-token generation limit; no custom stopping conditions. \\

Ashaar
& 0.55
& 0.90
& Not set
& 512
& 1.08
& Model EOS or the 512-token generation limit. Post-processing also truncates the output when decoded control tokens indicating meter or theme are encountered. \\

\bottomrule
\end{tabular}
\caption{Poem-generation decoding settings used for the evaluated systems. Top-$k$ was not explicitly supplied and therefore followed the relevant inference framework or model-configuration default.}
\label{tab:poem-generation-decoding}
\end{table*}

\section{Generation Parsing Details}
\label{app:generation-parsing}

Structural evaluation uses the \texttt{generated\_text} field for every system. We normalize line endings, remove invisible directional and control characters, trim surrounding whitespace, and remove empty lines. A line that already contains a recognized two-hemistich separator is treated as a complete bayt candidate. Otherwise, consecutive non-empty lines are paired in generation order: lines 1--2 form bayt 1, lines 3--4 form bayt 2, and so forth. Pairing is independent of the requested poem length.

With \texttt{REQUIRE\_TWO\_SHATR=True}, only complete two-hemistich bayts are classified. An unmatched final hemistich is retained only for coverage and audit accounting and is excluded from bayt-level structural classification.

\section{Meter Classifier}
\label{app:meter-classifier}

\begin{figure}[h]
    \centering
    \includegraphics[width=0.7\linewidth]{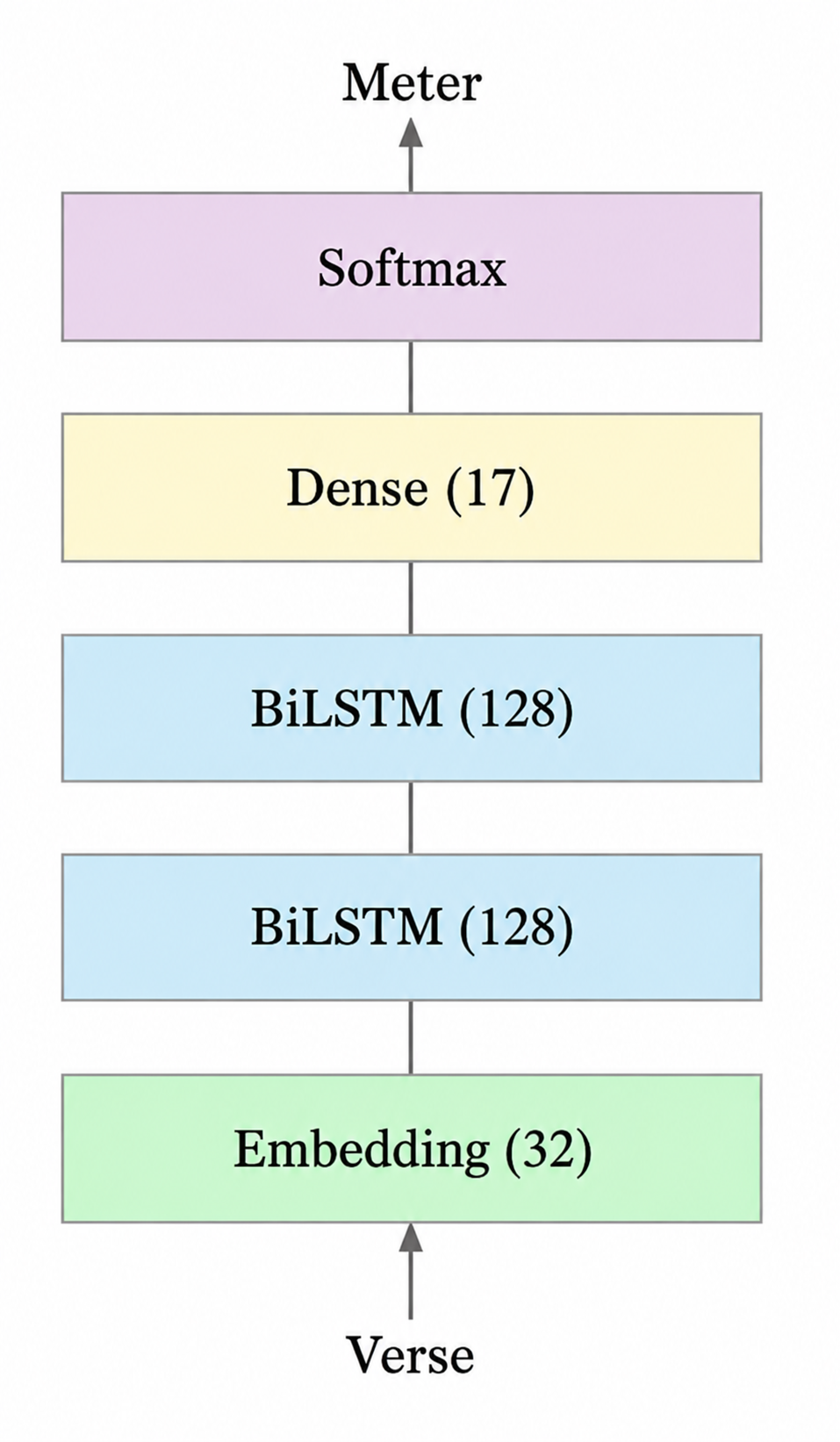}
    \caption{Architecture of the 4-BiLSTM Arabic meter classifier used for automatic meter evaluation.}
    \label{fig:4bilstm_architecture}
\end{figure}

The meter classifier follows the four-layer Bidirectional Long Short-Term Memory (BiLSTM) architecture of \citet{ABANDAH2022377}, shown in Figure~\ref{fig:4bilstm_architecture}. Each input verse is represented as a fixed-length character sequence ($T=128$) over a vocabulary of 46 characters, mapped into dense embeddings and processed by four stacked BiLSTM layers before a classification head predicts one of 17 classes corresponding to the supported meter labels and prose. Our PyTorch implementation is trained and evaluated on the \texttt{APCD\_plus\_prose\_all} dataset, achieving \textbf{97.24\%} accuracy on the held-out test split of 163,917 samples. The trained classifier is released as part of the project resources and is used for all metrical evaluations reported in this work.

\section{Meter-Subform Classifier}
\label{app:meter-subform-classifier}

Because the publicly available sub-meter checkpoint did not contain a complete trained sequence-classification head, we reconstructed the released AraPoemBERT sub-meter task \cite{qarah-2024-arapoembert} and trained an ARBERT-based classifier using the published configuration. Each complete bayt is represented as
\texttt{first\_hemistich S second\_hemistich}, with a maximum sequence length of 32 tokens.

The 102 unsupported test targets comprise Baseet--Majzu' (6 poems), Hazaj--Complete (18), Madeed--Complete (17), and Mujtath--Complete (61); Kamel--Ahadh is represented in the evaluator's class inventory.

\section{LLM-Based Evaluation}
\label{app:llm-judge}

Tables~\ref{tab:llm-judge-config} and~\ref{tab:api-judge-config} summarize the judge models and decoding configurations, while Table~\ref{tab:multi_judge_eval} reports the complete judge-specific results.

\begin{table}[h]
\centering
\footnotesize
\begin{tabular}{p{3.2cm}p{4.3cm}}
\toprule
\textbf{Parameter} & \textbf{Value} \\
\midrule
Judge model &
\makecell[l]{\texttt{qwen/}\\\texttt{qwen3-235b-a22b-2507}} \\
Temperature & 0.0 \\
Max output tokens & 128 \\
Reasoning effort & None \\
Tool calling & Forced \\
Output format & Integer score (1 to 5) \\
\bottomrule
\end{tabular}
\caption{Decoding configuration used for Qwen3-235B-A22B.}
\label{tab:llm-judge-config}
\end{table}

\begin{table*}[t]
\centering
\footnotesize
\begin{tabular}{p{3.2cm}p{5.4cm}p{5.4cm}}
\toprule
\textbf{Setting} & \textbf{Qwen 3.7 Max} & \textbf{GPT-5.6 Terra} \\
\midrule
Model/API identifier
& \texttt{qwen/qwen3.7-max}
& \texttt{openai/gpt-5.6-terra} \\

API provider
& OpenRouter Chat Completions
& OpenRouter Chat Completions \\

Temperature & 0 & 0 \\

Top-$p$ / top-$k$
& Provider defaults; fields not sent
& Provider defaults; fields not sent \\

Maximum output tokens & 128 & 128 \\

Reasoning effort & None & None \\

Seed
& Provider default; no seed sent
& Provider default; no seed sent \\

Structured output
& Forced \texttt{submit\_score} function call
& Forced \texttt{submit\_score} function call \\

Client-side fallback model & None & None \\
\bottomrule
\end{tabular}
\caption{API and decoding configurations used for Qwen 3.7 Max and GPT-5.6 Terra.}
\label{tab:api-judge-config}
\end{table*}

Each evaluation dimension is submitted through a separate request for every poem. The expected response is a forced function call containing an integer score from 1 to 5. Scores are parsed first from the function-call arguments. If these are unavailable, the parser accepts JSON content, fenced or embedded JSON, or a standalone integer from 1 to 5.

Malformed responses, parsing failures, and HTTP errors are retried up to three times after the initial attempt, allowing at most four attempts, with delays of 2, 4, and 6 seconds. The request timeout is 120 seconds, followed by a 0.25-second post-request interval. A permanently failed request terminates the run rather than silently excluding the corresponding evaluation. The completed runs contain no unresolved errors.

Qwen3-235B-A22B was applied to all 3,481 generated poems from each system, and its reported means are computed over the available valid scores. Qwen 3.7 Max and GPT-5.6 Terra evaluate the corresponding outputs from all four systems on a shared subset of 2,500 test instances, with complete coverage for every applicable system--dimension combination. \textsc{Shaer} and Yehia-7B are evaluated on all five dimensions, whereas Ashaar and Fanar-Diwan are evaluated on the four shared literary dimensions because their native interfaces do not support description conditioning.

For the cross-judge summary reported in the main text, scores are first averaged over the evaluated instances separately for each judge and system. The resulting three judge-level means are then averaged with equal weight, ensuring that each judge contributes equally despite the different evaluation-set sizes. All three judges use identical dimension-specific evaluation rubrics, provided below.

\begin{table*}[t]
\centering
\small
\begin{tabular}{llccccc}
\toprule
\textbf{System} & \textbf{Judge} & \textbf{Description} & \textbf{Meaning} & \textbf{Fluency} & \textbf{Coherence} & \textbf{Poeticness} \\
\midrule
\textsc{Shaer} & Qwen3-235B-A22B & 4.75 & 3.72 & 4.31 & 3.69 & 3.82 \\
Yehia-7B & Qwen3-235B-A22B & 4.42 & 3.75 & 4.21 & 3.82 & 3.73 \\
Fanar-Diwan & Qwen3-235B-A22B & -- & 3.70 & 4.26 & 3.69 & 3.79 \\
Ashaar & Qwen3-235B-A22B & -- & 2.84 & 3.46 & 2.84 & 3.00 \\
\midrule
\textsc{Shaer} & Qwen 3.7 Max & 3.88 & 2.35 & 2.07 & 2.37 & 2.19 \\
Yehia-7B & Qwen 3.7 Max & 3.39 & 2.56 & 2.23 & 2.64 & 2.13 \\
Fanar-Diwan & Qwen 3.7 Max & -- & 2.33 & 2.06 & 2.35 & 2.17 \\
Ashaar & Qwen 3.7 Max & -- & 2.05 & 2.01 & 2.18 & 1.98 \\
\midrule
\textsc{Shaer} & GPT-5.6 Terra & 3.75 & 2.19 & 2.15 & 2.49 & 2.21 \\
Yehia-7B & GPT-5.6 Terra & 3.30 & 2.43 & 2.50 & 2.81 & 2.09 \\
Fanar-Diwan & GPT-5.6 Terra & -- & 2.13 & 2.05 & 2.41 & 2.12 \\
Ashaar & GPT-5.6 Terra & -- & 1.82 & 2.06 & 2.08 & 1.80 \\
\bottomrule
\end{tabular}
\caption{Mean semantic and literary evaluation scores reported separately for each LLM judge. Qwen3-235B-A22B was applied to all 3,481 outputs per system, while Qwen 3.7 Max and GPT-5.6 Terra evaluate a shared subset of 2,500 test instances. Means are computed over the available valid scores. Description adherence applies only to \textsc{Shaer} and Yehia-7B; dashes indicate inapplicable evaluations.}
\label{tab:multi_judge_eval}
\end{table*}

\subsection{Evaluation Rubrics}

\subsubsection{Description Adherence}

\begin{itemize}
\item 1: Off-topic, contradictory, or ignores the central request.
\item 2: Touches only the broad topic while missing the intended meaning.
\item 3: Follows the main topic but only partially realizes the requested details.
\item 4: Clearly follows the description with only minor omissions.
\item 5: Precisely realizes the requested description and appears specifically written for that prompt.
\end{itemize}

\subsubsection{Meaning}

\begin{itemize}
\item 1: No recoverable meaning or severe contradictions.
\item 2: Weak meaning with little semantic substance.
\item 3: Understandable but ordinary or weakly developed meaning.
\item 4: Clear and worthwhile meaning with genuine semantic content.
\item 5: Rich and non-trivial poetic meaning exhibiting substantial depth or insight.
\end{itemize}

\subsubsection{Fluency}

\begin{itemize}
\item 1: Severely corrupted or ungrammatical Arabic.
\item 2: Frequent errors that substantially reduce readability.
\item 3: Generally understandable but containing noticeable awkwardness or errors.
\item 4: Mostly fluent with only minor imperfections.
\item 5: Consistently natural, correct, and elegant classical-style Arabic.
\end{itemize}

\subsubsection{Coherence}

\begin{itemize}
\item 1: Fragmented or internally inconsistent.
\item 2: Frequent topic shifts and weak overall unity.
\item 3: General thematic consistency but limited development.
\item 4: Mostly unified with clear line-to-line continuity.
\item 5: Strongly coherent with meaningful progression of ideas, imagery, and emotion.
\end{itemize}

\subsubsection{Poeticness}

\begin{itemize}
\item 1: Flat, mechanical, prose-like, or non-poetic output.
\item 2: Weak poetic quality with limited imagery or emotional impact.
\item 3: Some poetic characteristics but limited artistic development.
\item 4: Clearly poetic with effective literary qualities.
\item 5: Exceptional poetic quality with vivid imagery, strong emotional resonance, elegant rhetoric, and memorable diction.
\end{itemize}

\section{Human Evaluation}
\label{app:human-evaluation}

\subsection{Selection and Evaluation Procedure}

For ranking, the aggregate score of each candidate output is computed as the equal-weight mean across the three LLM judges and the four applicable evaluation dimensions. Selection is performed independently within each system.

System identities are removed, and the selected poems are presented in randomized order to reduce model-identity and presentation-order biases. Each annotator independently assigns one score per applicable dimension using the rubrics defined in Appendix~\ref{app:llm-judge}. Meter, meter-subform adherence, and length control are not assessed, as the human evaluation focuses exclusively on semantic and literary quality.

\subsection{Inter-Annotator Agreement}

We assess inter-annotator agreement separately for each dimension using ordinal Krippendorff's $\alpha$, accounting for the ordered 1--5 rating scale and the presence of three annotators.

Agreement is highest for Coherence ($\alpha=0.901$), while Meaning ($\alpha=0.758$), Fluency ($\alpha=0.751$), and Poeticness ($\alpha=0.756$) exhibit lower but similar agreement.

\section{Qualitative Examples from \textsc{Shaer}} \label{app:qualitative-examples} Table~\ref{tab:qualitative-examples} presents selected qualitative examples generated by \textsc{Shaer}. For each example, we report the input conditioning signals and the resulting meter conformity score.

\newcommand{\DescLabel}{\textenglish{\textbf{Description:}}}
\newcommand{\PoemLabel}{\textenglish{\textbf{Generated poem:}}}

\newcommand{\DescText}[1]{\begin{Arabic}\raggedleft\itshape #1\end{Arabic}}
\newcommand{\PoemHem}[1]{\begin{Arabic}\bfseries #1\end{Arabic}}

\begin{table*}[p]
\centering
\small
\setlength{\tabcolsep}{3pt}
\renewcommand{\arraystretch}{1.10}

\resizebox{0.96\textwidth}{!}{%
\begin{tabular}{
p{1.2cm}
p{1cm}
>{\centering\arraybackslash}p{1cm}
p{10.6cm}
>{\centering\arraybackslash}p{1.5cm}
}
\hline
\textbf{Meter} &
\textbf{Form} &
\textbf{Bayts} &
\textbf{Description and Generated Poem} &
\makecell[c]{\textbf{Meter}\\\textbf{score}} \\
\hline

\ArCell{البسيط} &
\ArCell{مخلع} &
\vspace{1pt}3 &
\begin{minipage}[t]{\linewidth}
\DescLabel{}

\vspace{2pt}
\DescText{القصيدة تتحدث عن شخص ظهرت عليه ملامح الحسن ، حيث تكشف ملامح وجهه عن فنون من الظرف والجمال، لدرجة أن وصله أصبح حقيقة يقينية تفوق ما كان في الخيال.}

\vspace{5pt}
\PoemLabel{}

\vspace{2pt}
\centering
\begin{tabular}{@{}r@{\hspace{1.5cm}}r@{}}
\PoemHem{للفضل في حسنه فنون} & \PoemHem{قد لاح لي شخصه فأبدى} \\
\PoemHem{وفي طلى طرفه فنون} & \PoemHem{في وجهه طالعات حسن} \\
\PoemHem{ما كان قبل الخيال ظن} & \PoemHem{قد صار وصلي به يقينا} \\
\end{tabular}
\end{minipage}
&
\vspace{1pt}0.9886 \\
\hline

\ArCell{الكامل} &
\ArCell{أحذ} &
\vspace{1pt}2 &
\begin{minipage}[t]{\linewidth}
\DescLabel{}

\vspace{2pt}
\DescText{القصيدة تتحدث عن لحظة ظهور عارضٍ لامعٍ في وجه الحبيب، يُقارن بجمال الريحان والآس، مما يدفع المتكلم إلى تقبيله فرحًا، لكن هذا التقبيل يتحول فجأة إلى سواد ناتج عن نيران أنفاسه، مما يعكس تباينًا بين الفرح والحرقة في مشهد عاطفي مكثف.}

\vspace{5pt}
\PoemLabel{}

\vspace{2pt}
\centering
\begin{tabular}{@{}r@{\hspace{1.5cm}}r@{}}
\PoemHem{فَكَأنَّهُ رَيحَانَةٌ وَآسُ} & \PoemHem{لاحَ العِذارُ بخَدِّهِ} \\
\PoemHem{فَغَدا السوادُ بِهِ هو المِقياسُ} & \PoemHem{قَبَّلْتُهُ طَرَباً وَلَهْواً} \\
\end{tabular}
\end{minipage}
&
\vspace{1pt}0.9859 \\
\hline

\ArCell{الرمل} &
\ArCell{مجزوء} &
\vspace{1pt}2 &
\begin{minipage}[t]{\linewidth}
\DescLabel{}

\vspace{2pt}
\DescText{القصيدة تتحدث عن تأثير الشيخوخة  الذي قوس القناتين بعد أن كانتا سويتين، وتصور الشعر الأبيض الذي يهتز برأس الشاعر وكأنه نشيد يهتف به الموت، مما يجعله عنوانًا للمنيّة.}

\vspace{5pt}
\PoemLabel{}

\vspace{2pt}
\centering
\begin{tabular}{@{}r@{\hspace{1.5cm}}r@{}}
\PoemHem{بعدَما كانت سويَّه} & \PoemHem{قوسَ الشيخُ قناتي} \\
\PoemHem{هتفت بي للمَنِيَّه} & \PoemHem{وإذا أبصرت شيبي} \\
\end{tabular}
\end{minipage}
&
\vspace{1pt}0.9941 \\
\hline

\ArCell{المجتث} &
\ArCell{تام} &
\vspace{1pt}2 &
\begin{minipage}[t]{\linewidth}
\DescLabel{}

\vspace{2pt}
\DescText{القصيدة تتحدث عن مفارقة كأس المحبة الذي يجمع بين المرارة والحلاوة، لتعبر عن طبيعة الحب الذي يبدو للوهلة الأولى بعيدًا لكنه في الحقيقة قريب جدًا من القلب.}

\vspace{5pt}
\PoemLabel{}

\vspace{2pt}
\centering
\begin{tabular}{@{}r@{\hspace{1.5cm}}r@{}}
\PoemHem{مرارة وحلاوة} & \PoemHem{كأس المحبة فيها} \\
\PoemHem{للقلب فيه غواية} & \PoemHem{والحب من بعد قرب} \\
\end{tabular}
\end{minipage}
&
\vspace{1pt}0.9627 \\
\hline

\ArCell{الرجز} &
\ArCell{تام} &
\vspace{1pt}2 &
\begin{minipage}[t]{\linewidth}
\DescLabel{}

\vspace{2pt}
\DescText{القصيدة تتحدث عن موقف الشجاعة والصدق في الخفاء، حيث يصف الشاعر نفسه بأنه يحمل ما حُمّل من خير وشر، ويؤكد أن صدق قناته لا يتشبه بوهج العذر الكاذب، مستخدمًا صورة الشهاب كالشرر في الكف لتدل على القوة الكامنة التي تظهر حين يختفي الجبان.}

\vspace{5pt}
\PoemLabel{}

\vspace{2pt}
\centering
\begin{tabular}{@{}r@{\hspace{1.5cm}}r@{}}
\PoemHem{حَمَّلْتُ ما حُمِّلْتُ من خيرٍ وشَرْ} & \PoemHem{إني إذا غبتَ وحُملْت الخبرْ} \\
\PoemHem{كأنَّها شُهُبٌ بكَفٍّ من شرَرْ} & \PoemHem{صدْقُ قناتي ليس كذباً كَوَهجْ} \\
\end{tabular}
\end{minipage}
&
\vspace{1pt}0.9968 \\
\hline

\ArCell{البسيط} &
\ArCell{مخلع} &
\vspace{1pt}2 &
\begin{minipage}[t]{\linewidth}
\DescLabel{}

\vspace{2pt}
\DescText{القصيدة تتحدث عن دعاء المولى أن يكون الله مفرجًا عن صدر الشاعر بعد أن طال به ضيقه واكتئابه، مستندًا في ذلك إلى فكرة أن اتساع الصدر يترافق مع النعيم، بينما ضيقه يترافق مع العذاب.}

\vspace{5pt}
\PoemLabel{}

\vspace{2pt}
\centering
\begin{tabular}{@{}r@{\hspace{1.5cm}}r@{}}
\PoemHem{ما طال بي الضِّيقُ والكَآبُ} & \PoemHem{يا ربِّ كن لي مُفَرِّجاً} \\
\PoemHem{والصَّدْرُ بالنَّغَماتِ ضائقُ} & \PoemHem{فالصدرُ بالنَّعماءِ واسِعٌ} \\
\end{tabular}
\end{minipage}
&
\vspace{1pt}0.6597 \\
\hline

\end{tabular}%
}

\caption{Qualitative examples generated by \textsc{Shaer}. For each example, the table reports the conditioning signals used for generation, the generated poem displayed in bayt structure, and its meter conformity score.}
\label{tab:qualitative-examples}
\end{table*}

\section{Performance Breakdowns}
\label{app:performance-breakdowns}

Tables~\ref{tab:base-meter-breakdown} and
\ref{tab:length-bucket-breakdown} report structural performance by
base meter and requested-length bucket, respectively. All 3,481
held-out instances are included for every system; no rows are excluded
from these breakdowns.

\begin{table*}[t]
\centering
\small
\setlength{\tabcolsep}{6pt}
\begin{tabular}{lrrrrr}
\toprule
\textbf{Base Meter} &
\textbf{$N$} &
\textbf{\textsc{Shaer}} &
\textbf{Yehia-7B} &
\textbf{Ashaar} &
\textbf{Fanar-Diwan} \\
\midrule
Baseet    & 529 & \textbf{98.49} & 24.76 & 95.27 & 79.21 \\
Khafeef   & 295 & 96.61 & 10.17 & \textbf{97.97} & 83.73 \\
Rajaz     & 137 & \textbf{80.29} & 15.33 & 64.23 & 58.39 \\
Ramal     & 138 & \textbf{84.06} & 9.42  & 83.33 & 63.04 \\
Sare'e    & 216 & \textbf{92.13} & 6.94  & 78.70 & 64.81 \\
Taweel    & 841 & \textbf{99.88} & 61.00 & 97.62 & 94.89 \\
Kamel     & 649 & \textbf{95.53} & 18.03 & 92.45 & 79.20 \\
Mutakareb & 156 & \textbf{96.79} & 4.49  & 91.03 & 79.49 \\
Mujtath   & 61  & \textbf{83.61} & 6.56  & 73.77 & 63.93 \\
Madeed    & 17  & \textbf{64.71} & 5.88  & 41.18 & 29.41 \\
Munsareh  & 84  & \textbf{82.14} & 0.00  & 75.00 & 48.81 \\
Hazaj     & 18  & \textbf{83.33} & 0.00  & 55.56 & 50.00 \\
Wafer     & 340 & \textbf{95.59} & 20.59 & 95.29 & 78.82 \\
\midrule
Overall   & 3,481 & \textbf{95.17} & 26.49 & 91.30 & 79.60 \\
\bottomrule
\end{tabular}
\caption{Poem-level base-meter accuracy (\%) by base meter. $N$ is
the number of held-out test poems associated with each meter. All
systems are evaluated on the same aligned test instances, and every
instance is included. The highest accuracy in each row is bolded.}
\label{tab:base-meter-breakdown}
\end{table*}

\begin{table*}[t]
\centering
\small
\setlength{\tabcolsep}{5pt}
\begin{tabular}{lrrrrrrr}
\toprule
&
&
\multicolumn{4}{c}{\textbf{Base-Meter Accuracy (\%)}} &
\multicolumn{2}{c}{\textbf{Count Accuracy (\%)}} \\
\cmidrule(lr){3-6}
\cmidrule(lr){7-8}
\textbf{Length Bucket} &
\textbf{$N$} &
\textbf{\textsc{Shaer}} &
\textbf{Yehia-7B} &
\textbf{Ashaar} &
\textbf{Fanar-Diwan} &
\textbf{\textsc{Shaer}} &
\textbf{Yehia-7B} \\
\midrule
1--3 verses   & 1,668 & \textbf{97.00} & 27.04 & 91.13 & 78.60 & \textbf{98.44} & 72.54 \\
4--6 verses   & 805   & \textbf{93.66} & 23.98 & 91.18 & 79.13 & \textbf{93.17} & 36.02 \\
7--10 verses  & 473   & \textbf{93.02} & 29.60 & 92.39 & 80.76 & \textbf{75.26} & 9.51 \\
11--20 verses & 535   & \textbf{93.64} & 25.79 & 91.03 & 82.43 & \textbf{28.97} & 0.56 \\
\midrule
Overall       & 3,481 & \textbf{95.17} & 26.49 & 91.30 & 79.60 & \textbf{83.40} & 44.47 \\
\bottomrule
\end{tabular}
\caption{Poem-level base-meter and exact count accuracy by requested
poem-length bucket. Count accuracy is reported only for \textsc{Shaer}
and Yehia-7B because Ashaar and Fanar-Diwan do not support requested
length conditioning. Every applicable held-out instance is included.}
\label{tab:length-bucket-breakdown}
\end{table*}

\end{document}